\documentclass[5p,times]{elsarticle}

\usepackage{hyperref}

\usepackage{amsmath}
\usepackage{amssymb}
\usepackage{multirow}
\usepackage{diagbox}
\usepackage{colortbl}
\usepackage{fixltx2e}

\usepackage[english]{babel}
\usepackage{blindtext}
\usepackage[utf8]{inputenc}
\usepackage[T1]{fontenc}
\usepackage{hyperref}
\usepackage{subcaption}
\usepackage{float}

\usepackage{xcolor}
\usepackage{multirow}
\usepackage{caption}
\usepackage{indentfirst}

\usepackage{xcolor}

\usepackage{placeins}

\newcommand*{\figuretitle}[1]{%
    {\centering
        \begin{footnotesize}
            \vspace{-.1em}#1\par\medskip\vspace{-.4em}
        \end{footnotesize}
    }
}

\makeatletter
\newcommand\footnoteref[1]{\protected@xdef\@thefnmark{\ref{#1}}\@footnotemark}
\makeatother

\journal{arXiv}

\begin{document}\sloppy

\begin{frontmatter}


\title{GCN-FFNN: A Two-Stream Deep Model for Learning Solution to Partial Differential Equations}

\author[]{Onur Bilgin}
\author[]{Thomas Vergutz}
\author[]{Siamak Mehrkanoon\corref{corauthor}}

\address{Department of Data Science and Knowledge Engineering, Maastricht University, The Netherlands}
\cortext[corauthor]{Corresponding author. S. Mehrkanoon is also with Mathematics Centre Maastricht, Maastricht University, The Netherlands \\
(e-mail: siamak.mehrkanoon@maastrichtuniversity.nl).}

\begin{abstract}
\indent

This paper introduces a novel two-stream deep model based on graph convolutional network (GCN) architecture and feed-forward neural networks (FFNN) for learning the solution of nonlinear partial differential equations (PDEs). The model aims at incorporating both graph and grid input representations using two streams corresponding to GCN and FFNN models, respectively. Each stream layer receives and processes its own input representation. As opposed to FFNN which receives a grid-like structure, the GCN stream layer operates on graph input data where the neighborhood information is incorporated through the adjacency matrix of the graph. 
In this way, the proposed GCN-FFNN model learns from two types of input representations, i.e. grid and graph data, obtained via the discretization of the PDE domain. The GCN-FFNN model is trained in two phases. In the first phase, the model parameters of each stream are trained separately. Both streams employ the same error function to adjust their parameters by enforcing the models to satisfy the given PDE as well as its initial and boundary conditions on grid or graph collocation (training) data. In the second phase, the learned parameters of two-stream layers are frozen and their learned representation solutions are fed to fully connected layers whose parameters are learned using the previously used error function. The learned GCN-FFNN model is tested on test data located both inside and outside the PDE domain. The obtained numerical results demonstrate the applicability and efficiency of the proposed GCN-FFNN model over individual GCN and FFNN models on 1D-Burgers, 1D-Schrödinger, 2D-Burgers and 2D-Schrödinger equations. 






\end{abstract}


\end{frontmatter}


\section{Introduction}

Partial differential equations (PDEs) are widely used in the mathematical formulation of physical phenomena in a variety of science and engineering applications such as modeling fluid flow, mechanical stress or material temperature among others. The analytic solutions of PDEs are not often available and therefore several numerical methods such as Finite Difference Methods (FDM) \cite{mohanty2005unconditionally}, Finite Element Methods \cite{thomee2001finite}, splines \cite{abushama2008modified, kumar2010methods}, finite volume method \cite{shakeri2011finite}, Spectral based method \cite{taleei2014time} have been developed for approximating the solution of the given PDEs. 

In particular, in finite difference-based methods, the domain of the PDE is discretized. The solution is only provided for the predefined grid points and additional interpolation is required to obtain the solution for the whole domain. Moreover, the method has a low accuracy in irregular domains, which limits its application in such domains. The finite-element method relies on the discretization of the domain via meshing which can be a challenging and time-consuming process, especially for complex geometries or higher-dimensional PDEs. In addition, similar to finite difference methods, the solution is approximated locally at each mesh point and therefore additional interpolation is required to find the solution at an arbitrary point in the domain \cite{mehrkanoon2015learning} 

Another class of methods that has been proposed in the literature for the simulation of dynamical systems is based on machine learning approaches and in particular kernel-based models as well as artificial neural networks. The use of neural network-based models for solving ordinary and partial differential equations goes back to the early '90s, see \cite{meade1994numerical, lee1990neural, van1995neural, ramuhalli2005finite}. The Hopfield neural networks are used in \cite{lee1990neural} to solve first-order differential equations. The authors in \cite{lagaris1998artificial} introduced a feed-forward neural network-based model to solve ordinary and partial differential equations. In their work, the model function is expressed as a sum of two terms where the first term, which contains no adjustable parameters, satisfies the initial/boundary conditions and the second term involves a trainable feed-forward neural network. In contrast to mesh-based approaches such as finite difference and finite element methods, neural network models (see \cite{meade1994numerical, choi2009comparison, shirvany2009multilayer}) can generate a closed-form solution and do not require meshing.


Mehrkanoon et al. \cite{mehrkanoon2012approximate, mehrkanoon2012ls, mehrkanoon2013ls, mehrkanoon2015learning}, for the first time, proposed a systematic machine learning approach based on primal-dual LS-SVM formulation to learn the solution of dynamical systems governed by a range of differential equations including ordinary differential equations (ODEs), partial differential equations (PDEs), differential algebraic equations (DAEs). Unlike the neural network based approaches described in \cite{lagaris1998artificial, lagaris2000neural} that the user had to define a form of a trial solution, which in some cases is not straightforward, in the LS-SVM based approach the optimal model is derived by incorporating the initial/boundary conditions as constraints of an optimization problem.

In particular, in LS-SVM based model \cite{mehrkanoon2012approximate, mehrkanoon2015learning}, the domain of the differential equation is first discretized to generate collocation points that are located inside the domain as well as on its initial and or boundary. Next, one starts with an LS-SVM representation of the solution in the primal and formulates a constrained optimization problem to obtain the optimal values for the model parameters (i.e. weights and biases). More precisely, the initial/boundary conditions of the differential equation are incorporated as constraints of an optimization problem. The aim of the formulated constrained optimization problem is to enforce the LS-SVM representation of the solution to satisfy the given differential equation on the collocation points inside the domain (through the defined objective of the optimization problem) as well as on the initial/boundary of the domain (through the defined hard constraints). One should note that this is not a regression task, as the solution of the differential equation is not provided during the training. In fact, by solving the constrained optimization problem, the optimal representation of the solution is obtained in the dual. The LSSVM-PDE code is available at \footnote{\url{https://github.com/SMehrkanoon/LSSVM-PDE-Solver}}.

It should also be noted that in the systematic machine learning approach presented in \cite{mehrkanoon2012approximate, mehrkanoon2012ls, mehrkanoon2015learning}, one can alternatively start with a different representation than the LS-SVM representation; for instance, a neural networks based representation, see \cite{raissi2017physics}. In addition, the hard constraints of the LS-SVM optimization formulation corresponding to the initial/boundary conditions of the PDE can also be relaxed and instead added as an additional term in the objective function of the optimization problem, see \cite{raissi2017physics}. Therefore, motivated by the systematic LS-SVM approach \cite{mehrkanoon2012approximate, mehrkanoon2015learning}, the authors in \cite{raissi2017physics} started with a feed-forward neural networks representation and introduced physics informed deep learning model and showed its effectiveness in solving differential equations. However, to the best of our knowledge, this existing link between the systematic LS-SVM approach \cite{mehrkanoon2012approximate, mehrkanoon2015learning} for solving differential equations and the physics informed deep learning model \cite{raissi2017physics} has not been explicitly stated in the literature. Sirignano and Spiliopoulos \cite{sirignano2018dgm} developed the Deep Galerkin Method (DGM), where the solution of high-dimensional PDE is approximated by a neural network. Zhu et al. \cite{zhu2019physics} developed a dense convolutional encoder-decoder network and E and Yu \cite{yu2017deep} proposed the Deep Ritz method, based on fully-connected layers and residual connections for solving PDEs.


It is the purpose of this paper to introduce a novel two-stream deep model based on Graph Convolutional Networks (GCNs) and feed-forward neural networks (FFNN) to learn the solution of the given differential equations. GCNs have been successfully applied in many application domains such as natural language processing \cite{peng2018large,zhang2019long}, computer vision \cite{johnson2018image,litany2018deformable} and weather elements forecasting \cite{stanczyk2021deep, aykas2021multistream} tasks.  The use of GCNs for learning the solution of PDEs has not been well explored. In the context of PDEs, the grid and graph input data are obtained by discretizing the PDE domain. The aim of the proposed model is to learn from both types of input representations, i.e. grid and graph data. 
In particular, FFNN processes the grid data, while GCNs operates on graph data and learns the relation between the features by incorporating the neighborhood information through the adjacency matrix of the graph. 
This paper is organized as follows. 
The proposed model is described in section \ref{Methodology}. The
numerical experiments and discussion of the obtained results are given in section \ref{Experiments}. The conclusion is drawn in section \ref{Conclusion}.

\section{Proposed Model}\label{Methodology}

\subsection{Graph Structure Data}

\begin{figure}[hbt!]
\center{}
  \begin{subfigure}{.44\linewidth}
    \includegraphics[width=\textwidth]{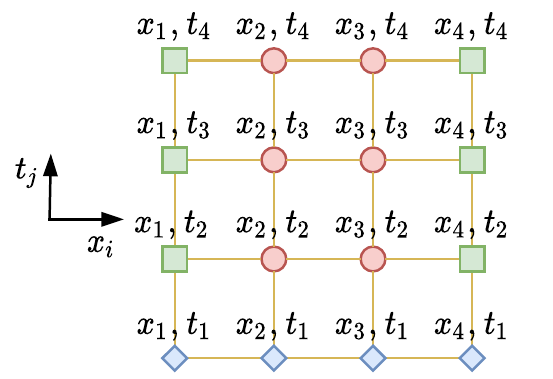}
    \caption{}
    \label{fig:graph_1}
  \end{subfigure}
  \begin{subfigure}{.54\linewidth}
    \includegraphics[width=\textwidth]{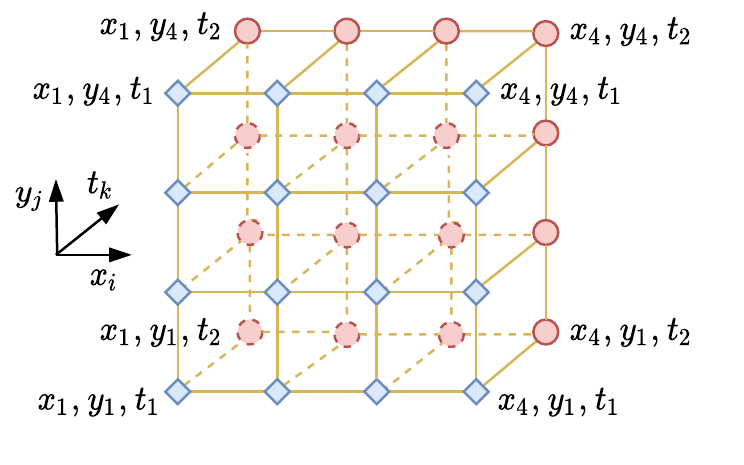}
    \caption{}
    \label{fig:graph_2}
  \end{subfigure}
  \caption{The used graph structures. (a) nodes with two attributes. (b) nodes with three attributes. Blue nodes are initial conditions and green nodes are boundary conditions. The remaining nodes inside the domain are depicted by red.}
  \label{fig:graph_structure}
\end{figure}

The domain of the PDE is first discretized into $N$ nodes. For a 2D-dimensional domain,  $N=X\times T$, where $X$ and $T$ are the numbers of elements in space and time dimensions, respectively. We next construct a graph where in particular the neighbors of the $(i,j)$-th node located inside the domain are $(i-1,j),(i+1,j),(i,j-1),(i,j+1)$-th nodes. The constructed graphs for equations with two and three-dimensional domains are shown in Fig. \ref{fig:graph_structure} (a) and (b), respectively. Here, in the case of two variables each node in the graph has neighbors in two directions $x$ and $t$, while in the case of three variables nodes have neighbors in three directions $x$, $y$ and $t$. In the constructed graph, let $Z_\mathcal{B}$ be the set of all boundary condition nodes, $Z_\mathcal{I}$ be the set of all initial condition nodes and $Z_\mathcal{BI}=Z_\mathcal{B} \cup Z_\mathcal{I}$. In addition, let $Z_D$ be the set of all nodes except the boundary and initial condition nodes, i.e. all nodes located inside the domain. The cardinality of the above-defined sets $Z_\mathcal{BI}$ and $Z_\mathcal{D}$ are denoted by $\lvert Z_\mathcal{BI} \rvert$ and $\lvert Z_\mathcal{D} \rvert$.


\subsection{Grid Structure Data}
The same discretization steps that were previously used to create the graph data are also used here to generate the grid data points. However, as opposed to previously introduced graph data, the grid data points do not have edge information and only contain the grid data points. 

\begin{figure*}[!hbt]
  \centering
  \includegraphics[width=0.8\linewidth]{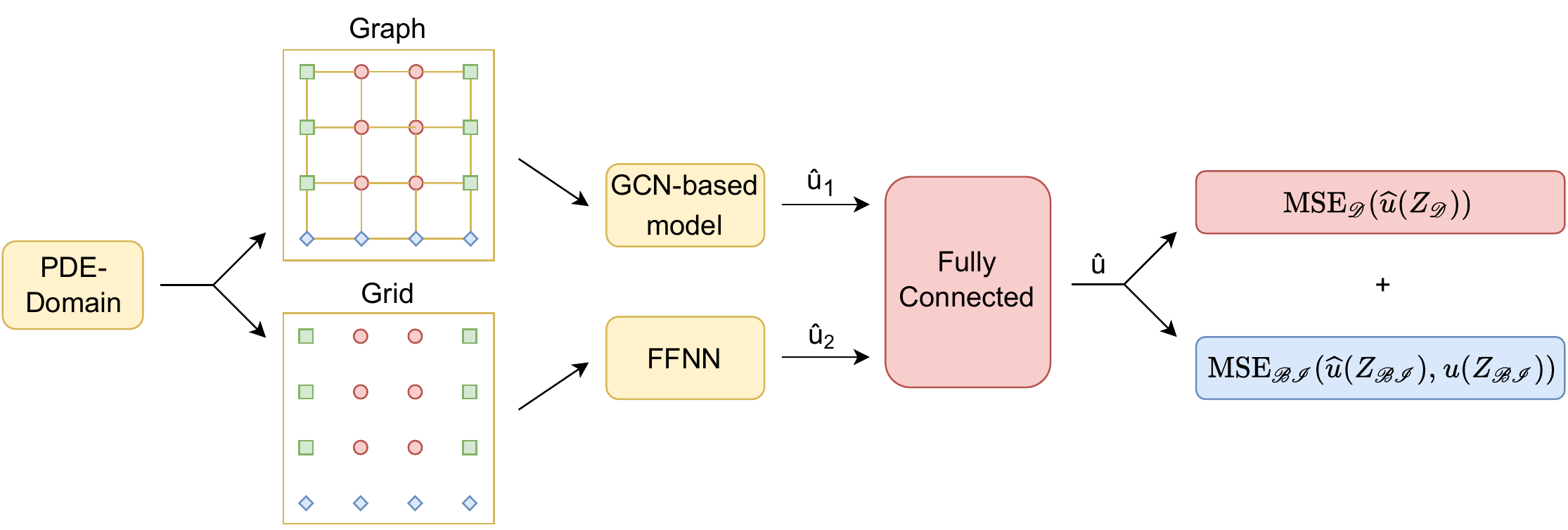}
  \caption{Architecture of the GCN-FFNN model for learning solutions to PDEs.}
  \label{fig:ensemble}
\end{figure*}

\subsection{GCN-FFNN model}
Here we propose our two-stream deep neural networks architecture which consists of Graph Convolutional Networks (GCNs) and feed-forward neural networks (FFNN) models in its first and second streams, respectively. The GCN-based model received the graph input data while the grid data are fed to the FFNN model. The architecture of the proposed GCN-FFNN model is shown in Fig. \ref{fig:ensemble}. The proposed model is trained in two phases. In the first phase, the models in the two streams are trained separately. The same error function is used to adjust the parameters of both streams by enforcing the models to satisfy the given PDE as well as its initial and boundary conditions on grid or graph training data. In the second phase, the learned parameters of two-stream layers are frozen and their learned representation solutions are fed
to fully connected layers whose parameters are learned using the previously employed error function. 
In addition to evaluating the proposed GCN-FFNN model, we have also individually examined each stream, i.e. FFNN as well as GCN-based models. It should be noted that the FFNN-based model has been previously proposed in the literature \cite{raissi2017physics}, whereas the GCN-based model and its combination with the FFNN-based model are introduced in this paper. 
In what follows, the GCN-based model in the GCN-FFNN architecture is explained in more detail.

\subsection{GCN-based model}\label{Graph Convolutional Network}
A model based on core Graph Convolutional Network (GCN) \cite{kipf2016semi} is developed and used in the first stream of the proposed GCN-FFNN model to learn the solution of partial differential equations. GCN is an efficient variant of convolutional neural networks which operates directly on graphs. In particular, the following layer-wise propagation rule is utilized in a multi-layer Graph Convolutional Network \cite{kipf2016semi}:

\begin{equation} \label{eq:3}
H^{(\ell+1)} = \sigma(\tilde{D}^{-\frac{1}{2}}\tilde{A}\tilde{D}^{-\frac{1}{2}}H^{(\ell)}W^{(\ell)}).
\end{equation}

\begin{figure}[!ht]
  \centering
  \includegraphics[width=0.9\linewidth]{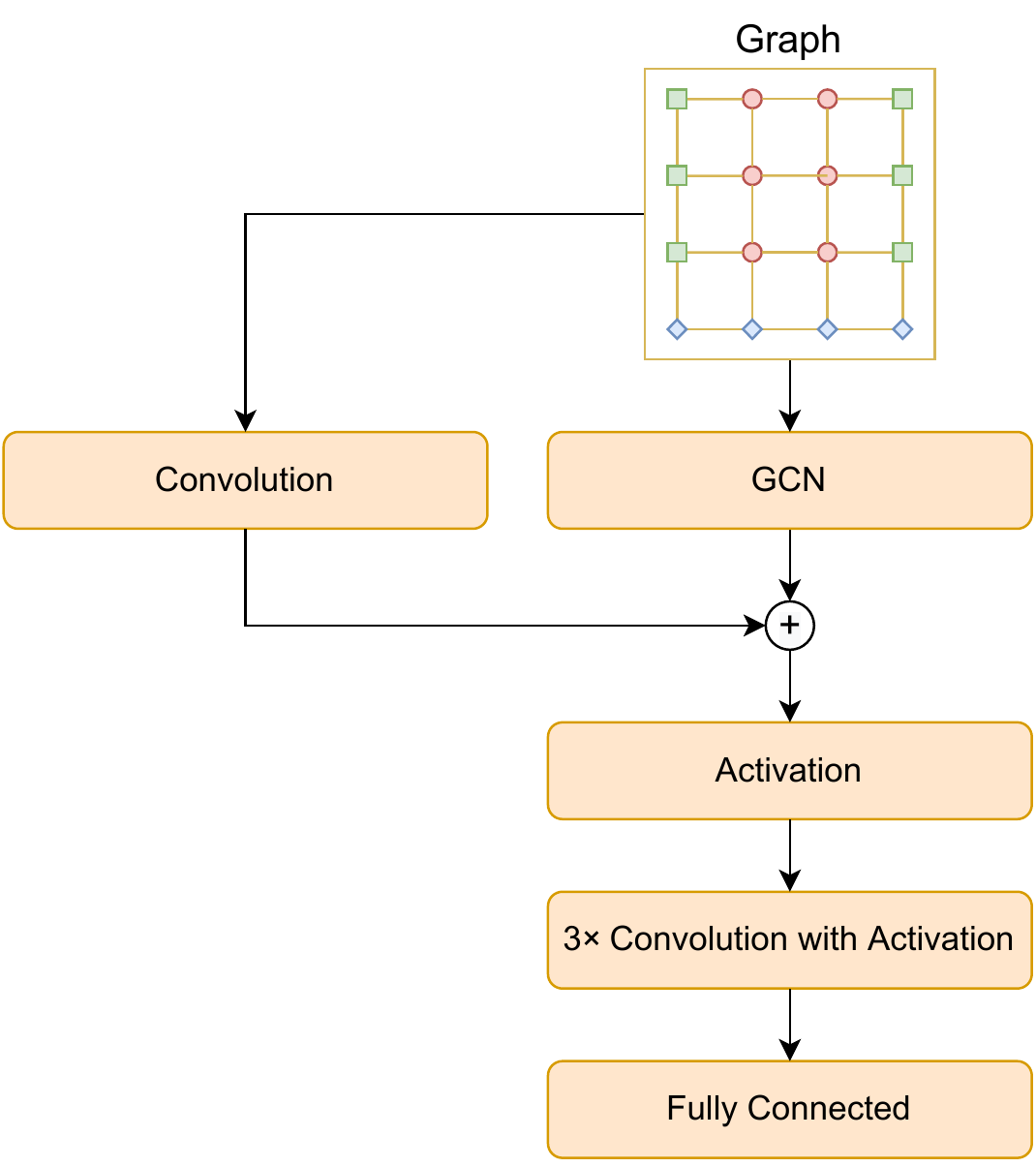}
  \caption{Architecture of the GCN-based model}
  \label{fig:model}
\end{figure}

Here, $\tilde{A}$ is the sum of adjacency matrix and the identity matrix to include self connections. $\tilde{D}=\textrm{diag}(d_1,d_2,\ldots,d_N)$ is the diagonal degree matrix with $d_i=\sum_{j}\tilde{A}_{ij}$. $H^{(\ell)}$ is the matrix of activations in the $\ell$-th layer with $H^{(0)}$ equals to the feature representation of the nodes. $W^{(\ell)}$ is the convolution weights for the $\ell$-th layer and $\sigma$ is the activation function, in our case the hyperbolic tangent activation function is used. The model receives the input of the shape $N \times P$, where $N$ is number of nodes and $P$ denotes the number of node attributes. In our case, for 2-dimensional PDE, $P=2$ and for 3-dimensional PDE, $P=3$. Furthermore, the graph structure, i.e. edge information, is also provided to the model through the adjacency matrix. The architecture of our proposed GCN-based model for learning the solutions of PDEs is shown in Fig. \ref{fig:model}. 
In particular, the model consists of one GCN layer followed by a residual connection and a hyperbolic tangent activation function. Next, three convolution layers with a filter size of $1 \times 1$, each with a hyperbolic tangent activation function are applied. The output of the last convolution layer is followed by a fully-connected layer. 

We use the systematic machine learning approach presented in \cite{mehrkanoon2015learning} to learn the parameters of the GCN model. The proposed GCN-based model receives all the nodes from $Z_\mathcal{BI}$ and $Z_\mathcal{D}$ sets as input as well as the partial differential operator $f[\cdot]$ which is obtained by putting all the involved terms of the given PDEs on the left-hand side of the equation so that the right-hand side of the PDE is zero. The model then outputs an approximate solution $\hat{u}$ for the given PDE. The solution $\hat{u}$ is learned by solving an optimization problem whose objective consists of two terms corresponding to the losses defined on inside domain nodes as well as on the initial/boundary condition nodes. The input data to the model consists of the above-mentioned 2- or 3-dimensional nodes and their graph structure (adjacency matrix). It should be noted that here the adjacency matrix is sparse due to the way that the edges of the graph are constructed, see section 2.1. Here, we use an efficient optimization code available at \cite{poli2019graph} for dealing with large scale sparse adjacency matrix in our GCN-layer.

\subsection{Loss Function}\label{Loss Function}

Following the work of Mehrkanoon and Suykens \cite{mehrkanoon2015learning}, here we use a loss function that enforces the representation of the solution, obtained by our proposed GCN-based model architecture, to satisfy the given differential equations and its initial/boundary conditions. Similar to \cite{mehrkanoon2015learning}, we aim at minimizing the mean squared loss function to adjust the model parameters. Here, the used loss function is composed of two terms given in Eq. (\ref{eq:3-3}). The first term, i.e. $\text{MSE}_{\mathcal{D}}$, enforces the representation of the solution to satisfy the given differential equation inside its domain. The second term, i.e. 
$\text{MSE}_{\mathcal{BI}}$, corresponds to making the difference between the true initial/boundary solutions and the model predictions as small as possible. More precisely, given the differential operator $f[\cdot]$ and the collocation nodes from both inside the domain (i.e. $Z_\mathcal{D}^{\,i}$) as well as on the initial/boundary of the domain (i.e. $Z_\mathcal{BI}^{\,i}$), the $\text{MSE}_{\mathcal{D}}$ and $\text{MSE}_{\mathcal{BI}}$ losses are defined in equations (\ref{eq:3-4}) and (\ref{eq:3-5}), respectively. 

\begin{equation} \label{eq:3-3}
\text{MSE}_\text{total}=\text{MSE}_{\mathcal{D}} + \text{MSE}_{\mathcal{BC}},
\end{equation}

\begin{equation} \label{eq:3-4}
\text{MSE}_{\mathcal{D}}=\frac{\sum_{i=1}^{\lvert Z_\mathcal{D} \rvert} f[\hat{u}](Z_\mathcal{D}^{\,i})}{\lvert Z_\mathcal{D} \rvert}.
\end{equation}

\begin{equation} \label{eq:3-5}
\text{MSE}_{\mathcal{BI}}=\frac{\sum_{i=1}^{\lvert Z_\mathcal{BI} \rvert} \large(u(Z_\mathcal{BI}^{\,i})-\hat{u}(Z_\mathcal{BI}^{\,i})\large)^2}{\lvert Z_\mathcal{BI} \rvert}.
\end{equation}

\section{Numerical Results}\label{Experiments}

\begin{figure}[hbt!]
\center{}
  \begin{subfigure}{.5\linewidth}
    \includegraphics[width=\textwidth]{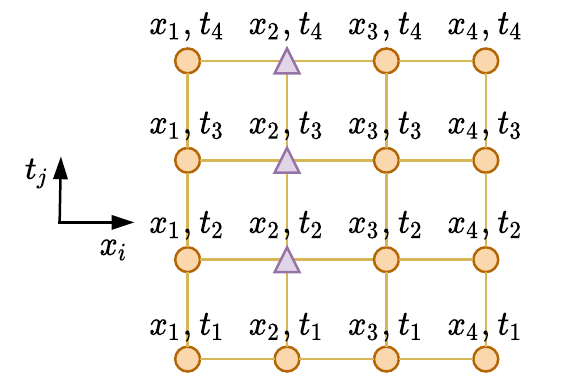}
    \caption{}
    \label{fig:test_samples_1}
  \end{subfigure}
  \begin{subfigure}{.5\linewidth}
    \includegraphics[width=\textwidth]{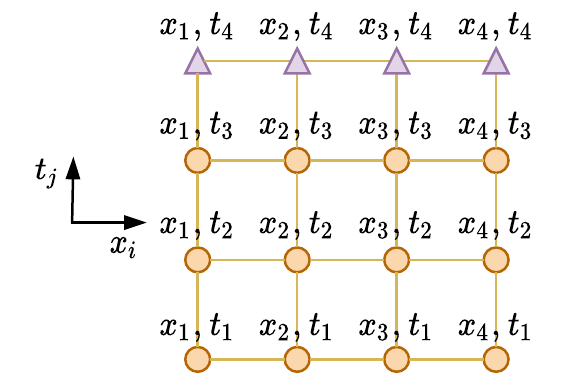}
    \caption{}
    \label{fig:test_samples_2}
  \end{subfigure}
  \caption{Two scenarios for selecting the test nodes. Purple triangle nodes are test samples and yellow circle nodes are training samples. (a): The test nodes are from inside the domain. (b): The test nodes are from outside the domain.}
  \label{fig:test_samples}
\end{figure}

In this section, four experiments are performed to demonstrate
the efficiency of the proposed GCN-FFNN model for learning the solution of 1D- and 2D-Burgers equations as well as 1D- and 2D-Schrödinger equations. The model parameters are learned in a transductive fashion. The input dataset is divided into train and test sets. The training set contains nodes from both inside the domain as well as its initial and boundary. Two scenarios are examined for test nodes, i.e. test nodes inside and outside the domain of PDE, see Fig. \ref{fig:test_samples}. In the first case, see Fig. \ref{fig:test_samples} (a), some grid points along the x-dimension are first randomly selected and then all the (x,t)-nodes with those selected x-coordinate positions form the inside domain test nodes. It should be noted that for PDEs with 3-dimensional domains, the random grid points are selected along the x- and y-dimensions and subsequently all the nodes (x,y,t)-nodes with those selected (x,y)-coordinate positions form the inside domain test nodes. In the second case, see Fig. \ref{fig:test_samples} (b), the test nodes are from outside the domain. In both test cases, the test set is 10\% of the whole dataset. 

The accuracy of the obtained approximate solution is measured by means of mean squared as well as infinite error norms defined as follows:

\begin{equation} \label{eq:test_loss}
\text{MSE}_\text{test}=\frac{\sum_{i=1}^{N_{\text{test}}} e_i^2}{N_\text{test}},\;\;\; \text{L}_\infty=||e||_\infty.
\end{equation}
Here, $e_i=u(Z^i_{\text{test}})-\hat{u}(Z^i_{\text{test}})$ where $Z^i_{\text{test}}$ is the $i$-th test node and $N_\text{test}$ is the number of test nodes. The obtained results of three models, i.e. FFNN, GCN and GCN-FFNN, on the above-mentioned four equations are compared. All the models are trained using L-BFGS with a learning rate of 1.0 for a maximum of 50000 epochs \cite{liu1989limited}. In order to make a fair comparison, all the models receive the same training and test sets. The number of layers, hidden neurons and trainable parameters of each of the used modules of GCN-FFNN architecture for each question are empirically found and tabulated in Table \ref{tab:hyper-parameters}.

\begin{table*}[hbt!]
\centering
\caption{The empirically found hyper-parameters of each module of the GCN-FFNN model for each equations.}
\label{tab:hyper-parameters}
\begin{tabular}{lcccc}
               PDE             &  Modul          &  \#  Layers  & \# Hidden Units  & \# Trainable Parameters \\

\hline
\multirow{3}{*}{1D-Burgers}  & FFNN-based model \cite{raissi2017physics}  &  8  & 20           & 2601 \\
                            & GCN-based model & 1 & 12  &   553   \\
                            & Fully-Connected & 2 & 48 & 144 \\
\hline
\multirow{3}{*}{1D-Schrödinger} & FFNN-based model \cite{li2021deep} &  6   & 100           & 40902   \\
                            & GCN-based model & 1 & 256  &     199426 \\
                            & Fully-Connected & 1 & 1  & 2 \\
\hline
\multirow{3}{*}{2D-Burgers} & FFNN-based model \cite{li2021deep}   & 8  & 20           & 2621   \\
                            & GCN-based model & 1 & 12  & 577 \\
                            & Fully-Connected & 2 & 16  &  48 \\
\hline
\multirow{3}{*}{2D-Schrödinger} & FFNN-based model \cite{li2021deep}  &  5   & 50           & 7952  \\
                            & GCN-based model & 1 & 18 & 1208  \\ 
                            & Fully-Connected & 2 & 16 & 48 \\
\hline
\end{tabular}
\end{table*}

\begin{table*}[hbt!]
\centering
\caption{The obtained MSEs and infinity norm errors for inside and outside test sets.}
\label{tab:loss-inside}
\begin{tabular}{l|c|cc|cc}
\multicolumn{1}{c|}{\multirow{3}{*}{PDE}} & \multirow{3}{*}{Model} & \multicolumn{4}{c}{Test nodes} \\ \cline{3-6} 
\multicolumn{1}{c|}{} &  & \multicolumn{2}{c|}{Inside the domain} & \multicolumn{2}{c}{Outside the domain} \\
\multicolumn{1}{c|}{} &  & $\text{MSE}_{\text{test}}$ & \multicolumn{1}{c|}{$\text{L}_\infty$} & $\text{MSE}_{\text{test}}$ & $\text{L}_\infty$ \\ \hline
\multirow{3}{*}{1D-Burgers}  & FFNN          & $5.10\cdot10^{-6}$    &  $0.025$   & $6.04\cdot10^{-6}$      &    $0.029$     \\
                            & GCN   & $6.44\cdot10^{-4}$       &      $0.139$      & $8.81\cdot10^{-4}$       &        $0.383$       \\
                            & GCN-FFNN     &\boldmath{$3.87\cdot10^{-6}$}      &    \boldmath{$0.022$}    & \boldmath{$1.50\cdot10^{-6}$}       &        \boldmath{$0.019$}       \\ 
\hline
\multirow{3}{*}{1D-Schrödinger} & FFNN        & \boldmath{$9.00\cdot10^{-6}$}              &    \boldmath{$0.008$}  & $5.42\cdot10^{-5}$         &   \boldmath{$0.017$}   \\
                            & GCN    & $1.30\cdot10^{-4}$     &    $0.023$ & $9.48\cdot10^{-4}$    &    $0.030$  \\
                            & GCN-FFNN     & $8.75\cdot10^{-5}$         &   $0.011$ & \boldmath{$3.39\cdot10^{-5}$}    &    $0.027$ \\ 
\hline
\multirow{3}{*}{2D-Burgers} & FFNN     & $1.68\cdot10^{-3}$ &  $0.085$       & $4.52\cdot10^{-3}$    &    $0.086$\\
                            & GCN    & $2.27\cdot10^{-3}$    &     $0.094$  & \boldmath{$5.92\cdot10^{-4}$}        &   $0.030$ \\
                            & GCN-FFNN         & \boldmath{$1.49\cdot10^{-3}$}    &    \boldmath{$0.077$}  & $5.99\cdot10^{-4}$        &   \boldmath{$0.027$} \\

\hline
\multirow{3}{*}{2D-Schrödinger} & FFNN      & \boldmath{$1.47\cdot10^{-7}$} &    \boldmath{$0.002$}  & $3.02\cdot10^{-7}$  &   \boldmath{$0.002$} \\
                            & GCN     & $1.19\cdot10^{-6}$      &  $0.003$  & $1.51\cdot10^{-6}$  &  $0.003$   \\ 
                            & GCN-FFNN    & \boldmath{$1.47\cdot10^{-7}$}  &   \boldmath{$0.002$}  & \boldmath{$2.58\cdot10^{-7}$}  &  \boldmath{$0.002$}     \\ 

\hline
\end{tabular}
\end{table*}

\subsection{1D-Burgers Equation}\label{Burgers Equation}

The 1D-Burgers equation with boundary and initial conditions is given in Eq. (\ref{eq:1d-burgers-1}) \cite{raissi2017physics}: 

\begin{equation} \label{eq:1d-burgers-1}
\begin{cases}
u_t+uu_x-(0.01/\pi)u_{xx}=0,\;\;\; x \in [-1,1],\;\;\; t\in [0,0.99], \\
u(x,0)=-sin(\pi x), \\
u(-1,t)=u(1,t)=0.
\end{cases}
\end{equation}
Here, the differential operator is defined as $f[u]:= u_t+uu_x-(0.01/\pi)u_{xx}$. $u(x,t)$ (shown as $u$ in the equation) is the solution we seek to approximate with our proposed model. $u_x$ and $u_t$ are the partial derivatives of $u$ with respect to $x$ and $t$, respectively. $u_{xx}$ is the second partial derivative of $u$ with respect to $x$. 

In the first scenario, the two dimensional domain of the 1D-Burgers equation, i.e. $(x,t) \in [-1,1]\times [0,0.99]$, is divided into $N=256\times100$ nodes, which are evenly spaced in each dimension. We have randomly selected $10\%$ of $N$ nodes to form the test nodes inside the domain. In the second scenario, the nodes in the ranges $(x,t) \in [-1,1]\times [0,0.89]$ are used for training and the test nodes outside the domain are selected from $0.89 < t\leq 0.99$. The obtained MSEs and infinity norm of each model are tabulated in Table \ref{tab:loss-inside}. Both used metrics show that the proposed GCN-FFNN model achieved the best results for both inside and outside test nodes. 

\begin{figure*}[hbt!]
\center{}
  \begin{subfigure}{.32\textwidth}
    \figuretitle{The plots of $u(x,t)$ at $x=-0.15$}
    \includegraphics[width=\textwidth]{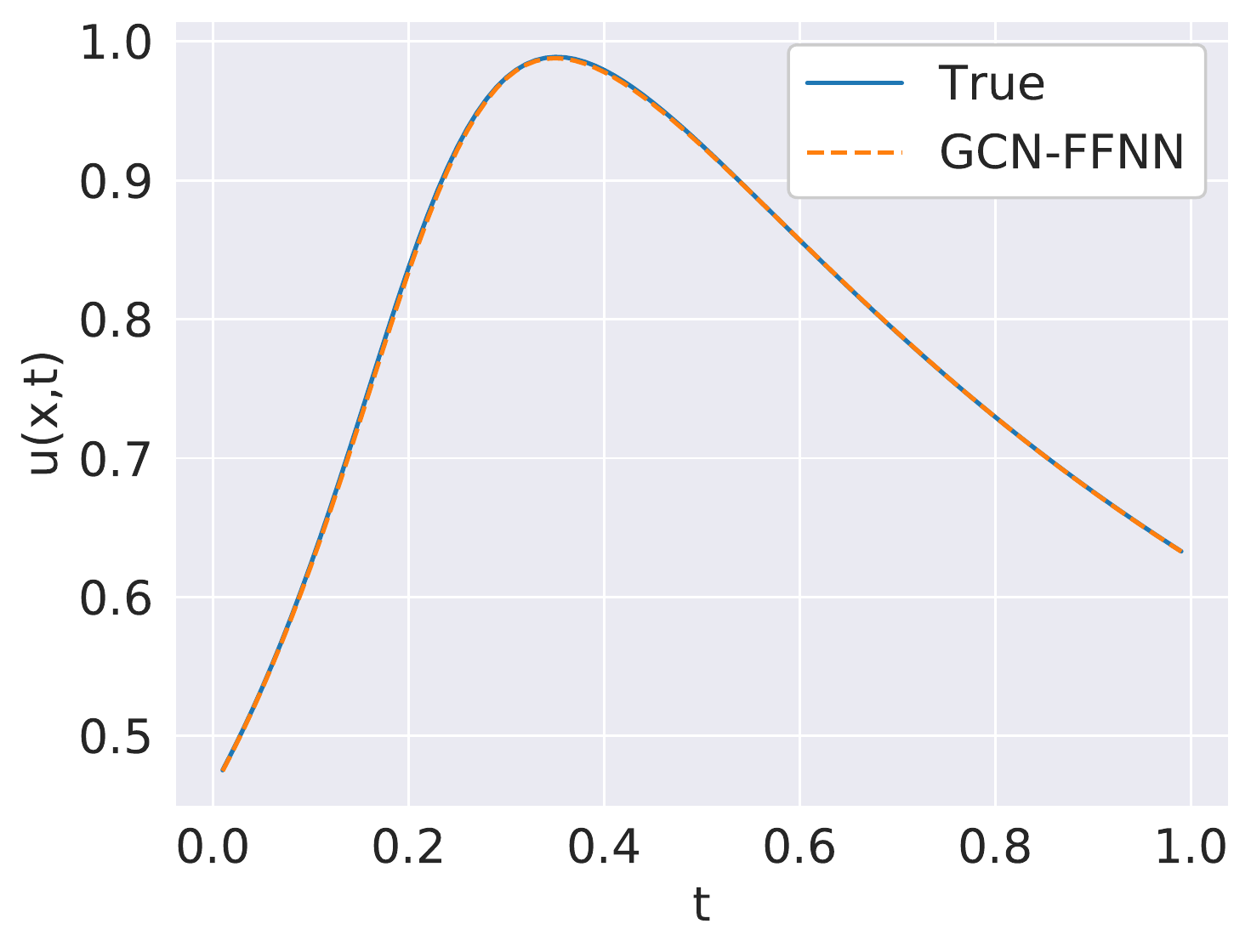}
    \vspace{-.11\textwidth}\caption{}\vspace{.05\textwidth}
    \label{fig:x1_inside_1D_burgers}
  \end{subfigure}
  \begin{subfigure}{.325\textwidth}
    \figuretitle{The plots of $u(x,t)$ at $x=0.15$}
    \includegraphics[width=\textwidth]{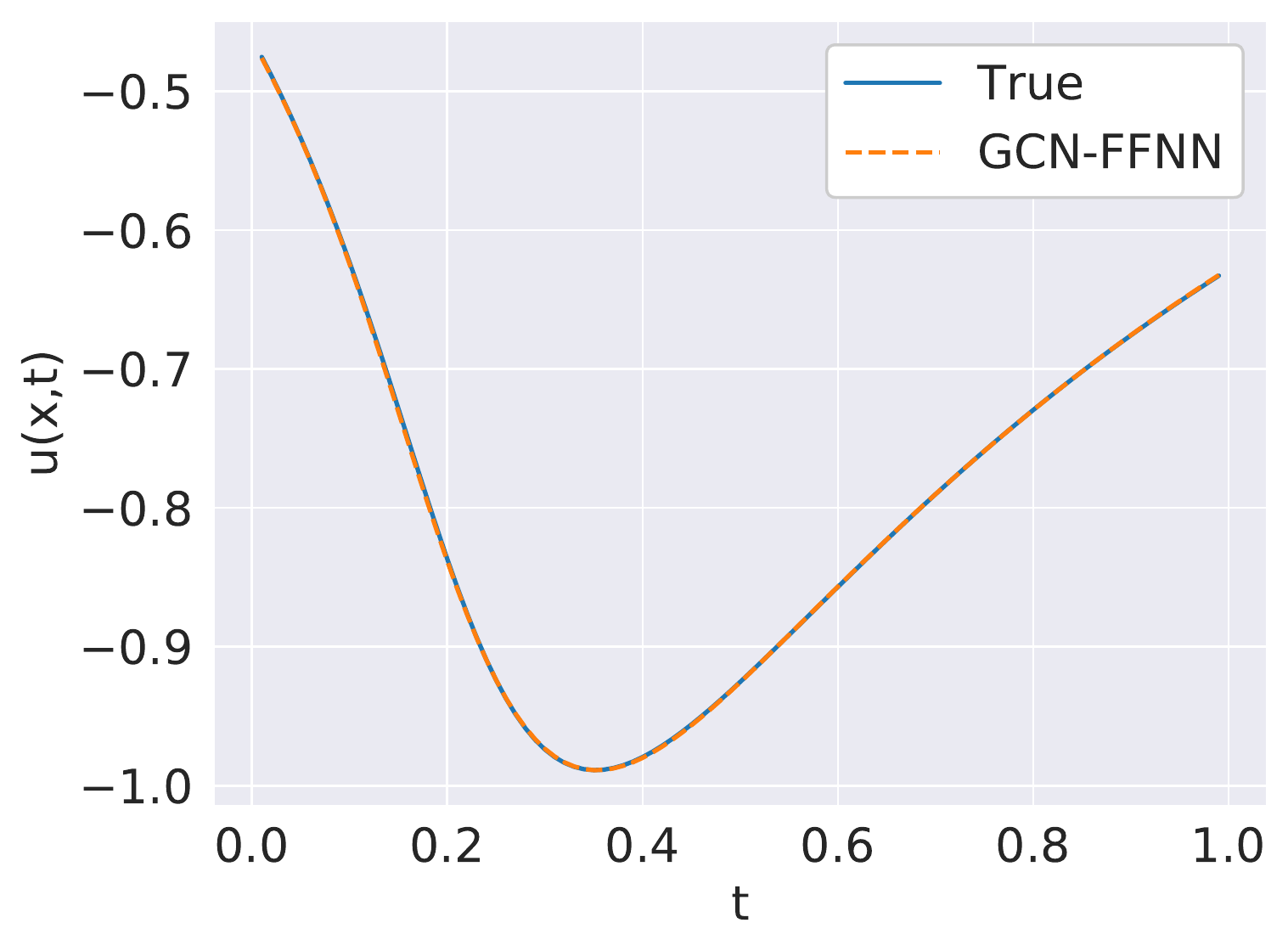}
    \vspace{-.11\textwidth}\caption{}\vspace{.05\textwidth}
    \label{fig:x2_inside_1D_burgers}
  \end{subfigure}
  \begin{subfigure}{.33\textwidth}
    \figuretitle{The plots of $u(x,t)$ at $x=0.94$}
    \includegraphics[width=\textwidth]{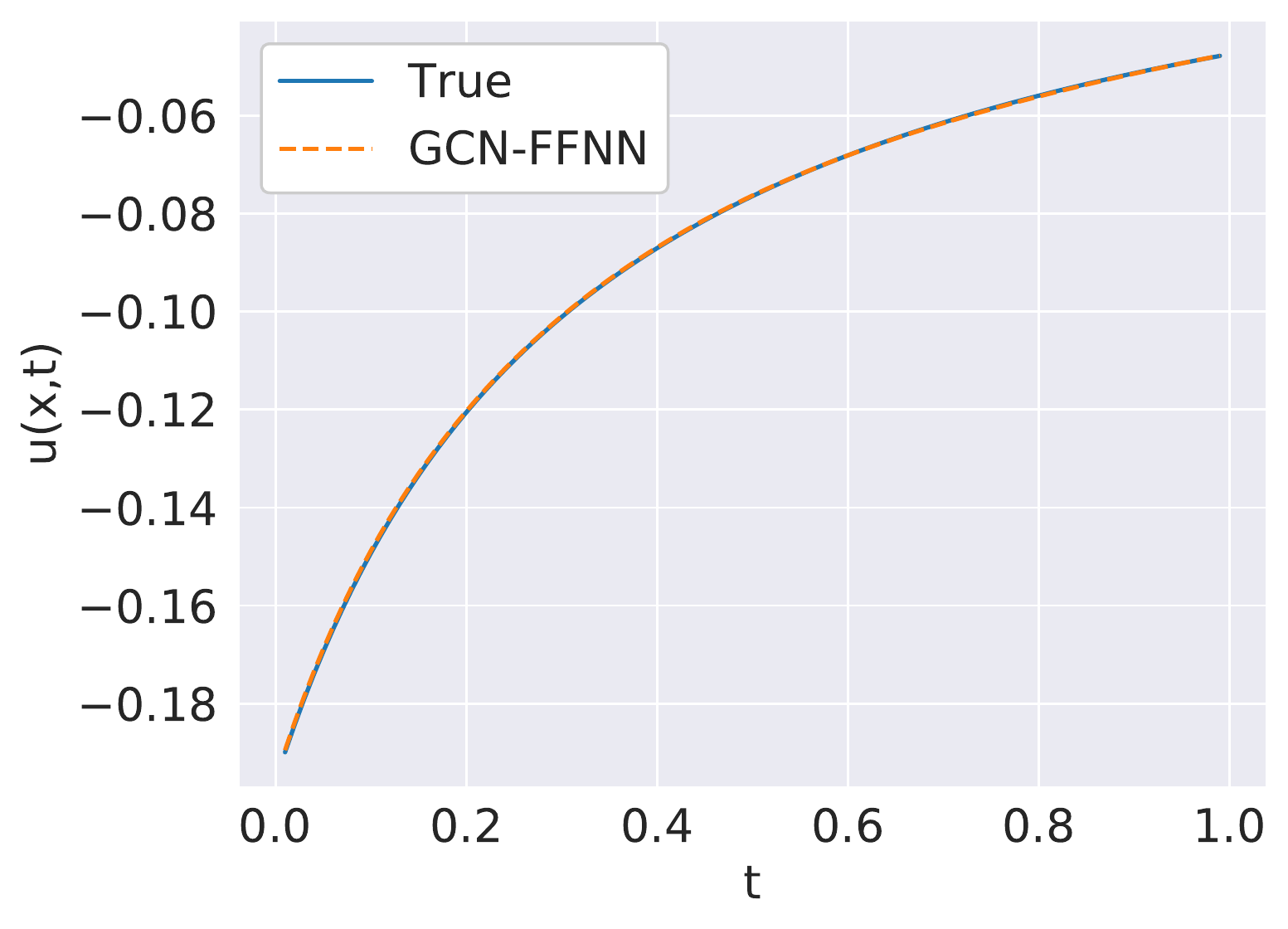}
    \vspace{-.11\textwidth}\caption{}\vspace{.05\textwidth}
    \label{fig:x3_inside_1D_burgers}
  \end{subfigure}
  \begin{subfigure}{.33\textwidth}
    \figuretitle{The plot of $u(x,t)-\hat{u}(x,t)$ at $x=-0.15$}
    \includegraphics[width=\textwidth]{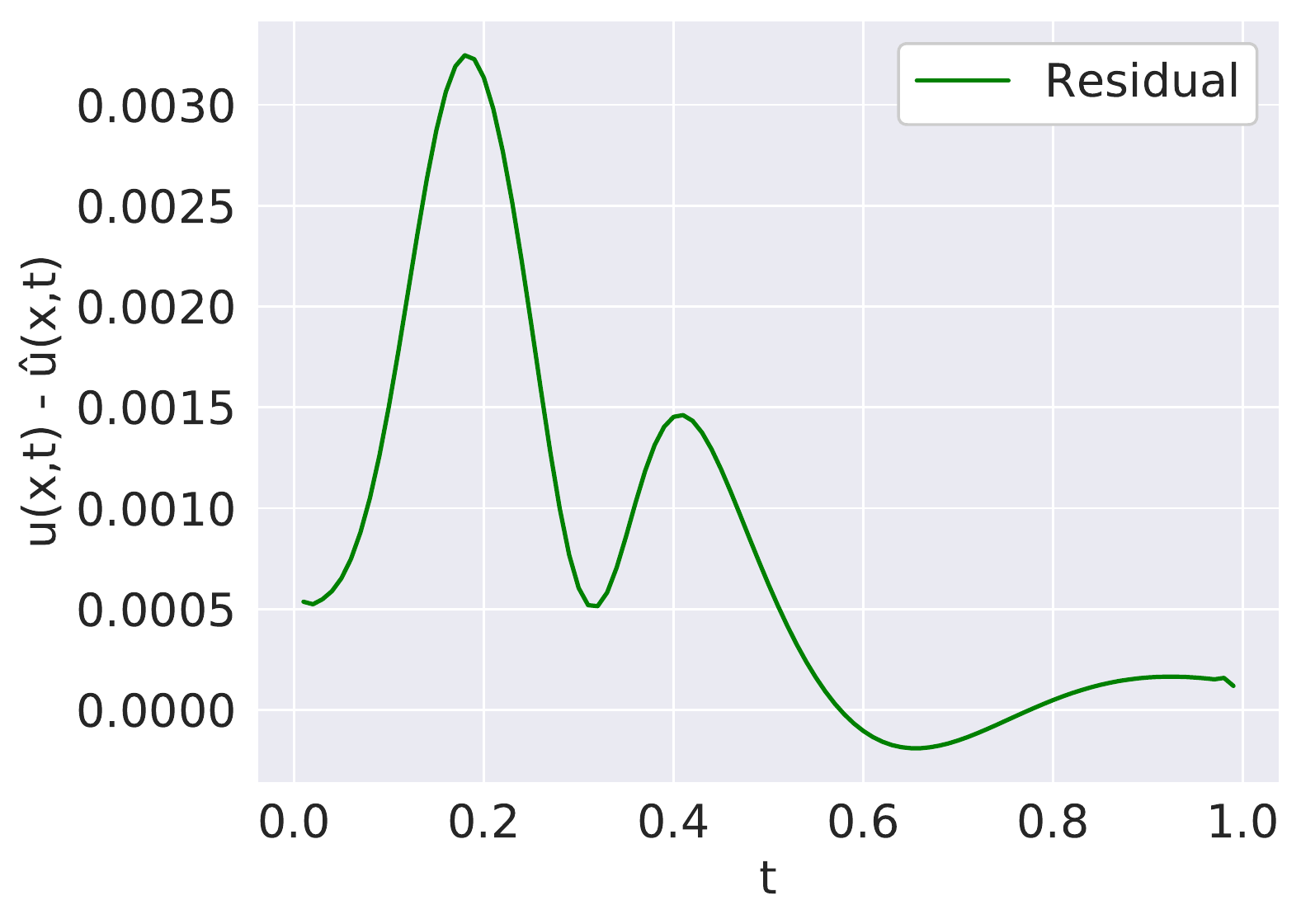}
    \vspace{-.11\textwidth}\caption{}
    \label{fig:r1_inside_1D_burgers}
  \end{subfigure}
  \begin{subfigure}{.33\textwidth}
    \figuretitle{The plot of $u(x,t)-\hat{u}(x,t)$ at $x=0.15$}
    \includegraphics[width=\textwidth]{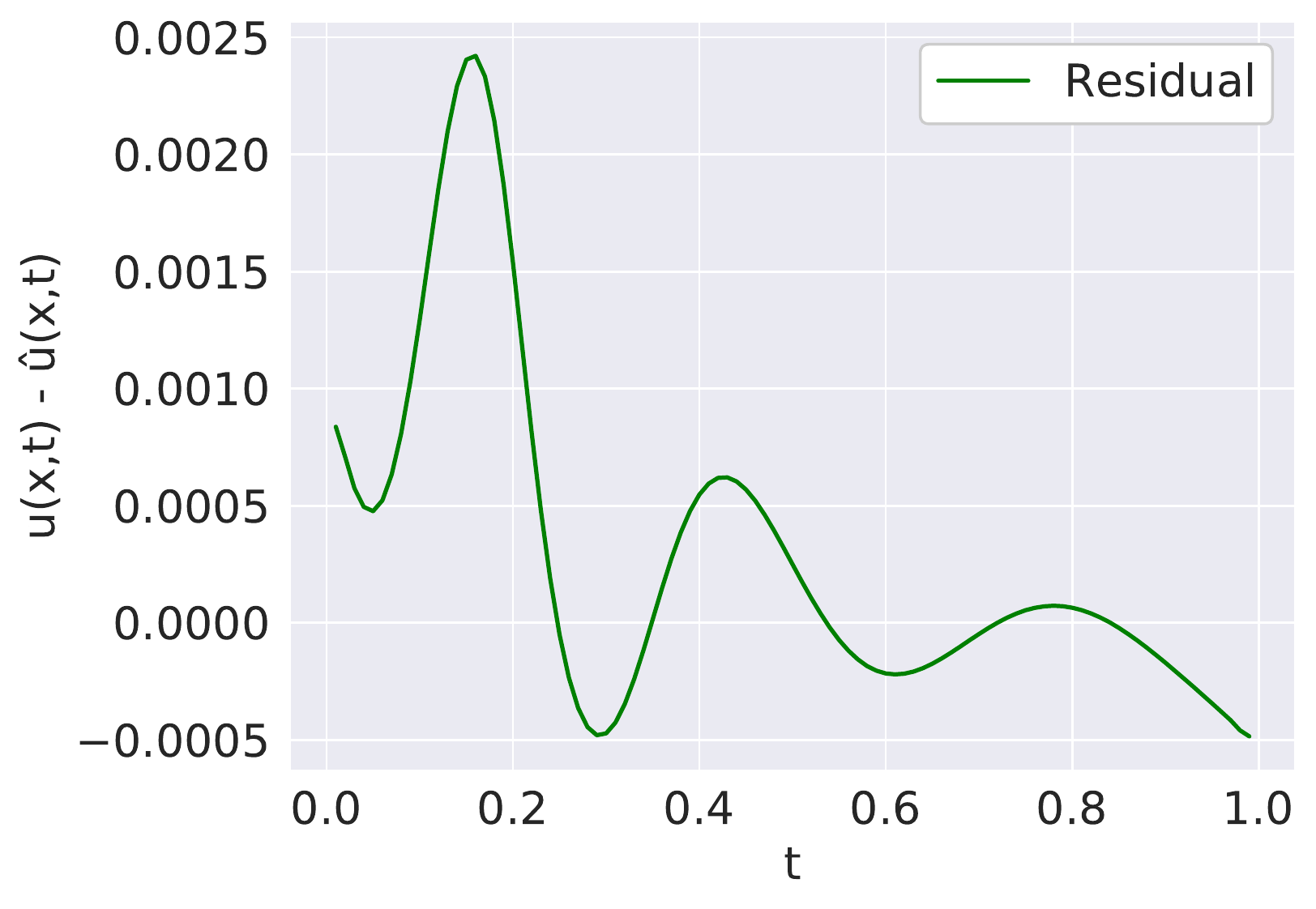}
    \vspace{-.11\textwidth}\caption{}
    \label{fig:r2_inside_1D_burgers}
  \end{subfigure}
  \begin{subfigure}{.33\textwidth}
    \figuretitle{The plot of $u(x,t)-\hat{u}(x,t)$ at $x=0.94$}
    \includegraphics[width=\textwidth]{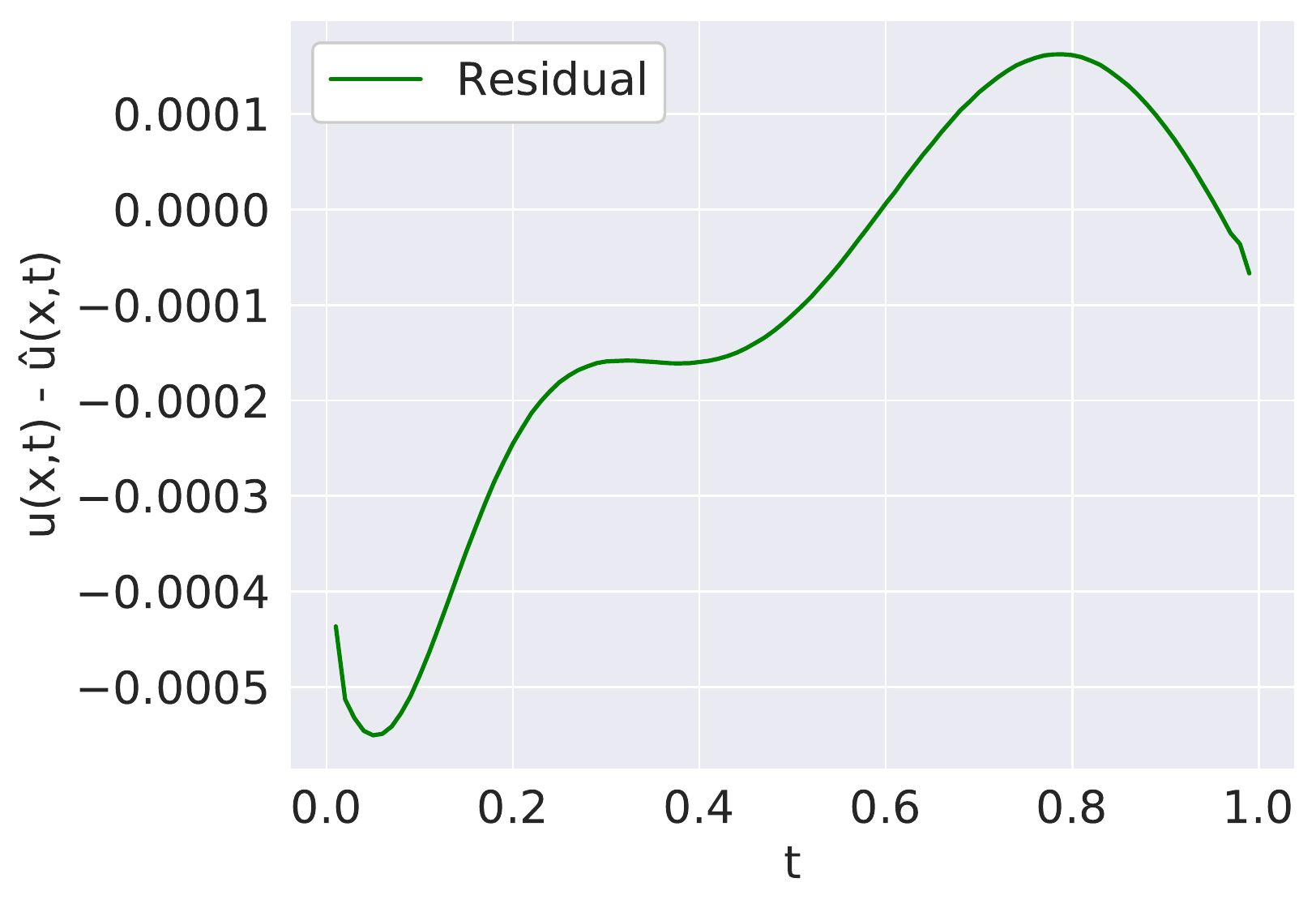}
    \vspace{-.11\textwidth}\caption{}
    \label{fig:r3_inside_1D_burgers}
  \end{subfigure}
  \caption{\textbf{First scenario:} The plots for the 1D-Burgers equation obtained by GCN-FFNN model corresponding to the first scenario where test nodes are from inside the domain. (a),(b) and (c): The true and approximate solution obtained by GCN-FFNN model. (d),(e) and (f): The obtained residuals $u(x,t)-\hat{u}(x,t)$. The data at $x=-0.15$ belongs to the training set, while the data at $x = 0.15$ and $x = 0.94$ are from test set.}
  \label{fig:1d-burgers_plots_inside}
\end{figure*}

\begin{figure*}[hbt!]
\center{}
  \begin{subfigure}{.3\textwidth}
    \figuretitle{The plots of $u(x,t)$ at $t=0.50$}
    \includegraphics[width=\textwidth]{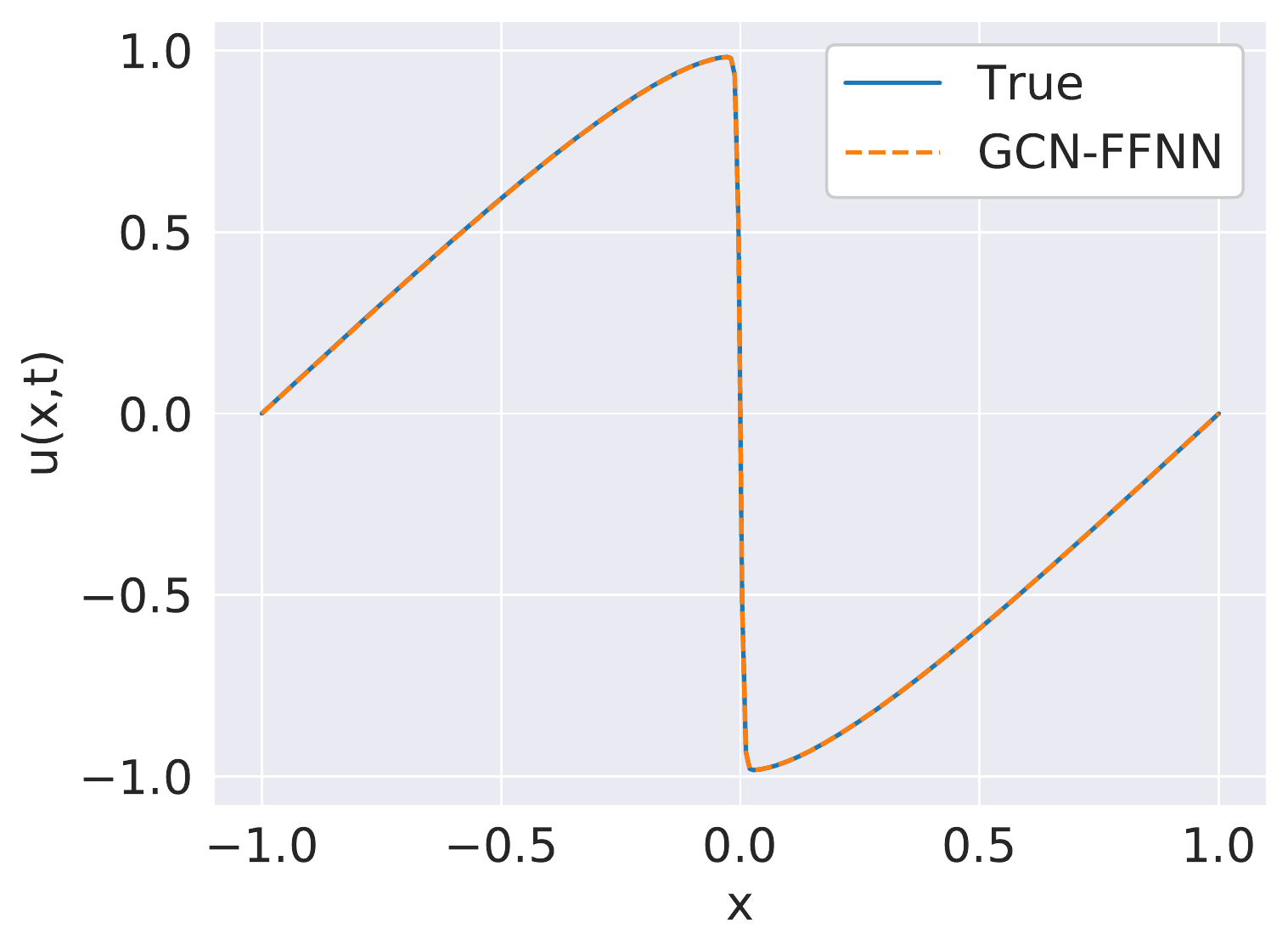}
    \vspace{-.11\textwidth}\caption{}\vspace{.05\textwidth}
    \label{fig:t050_outside_1D_burgers}
  \end{subfigure}\hspace{0.02\textwidth}
  \begin{subfigure}{.3\textwidth}
    \figuretitle{The plots of $u(x,t)$ at $t=0.75$}
    \includegraphics[width=\textwidth]{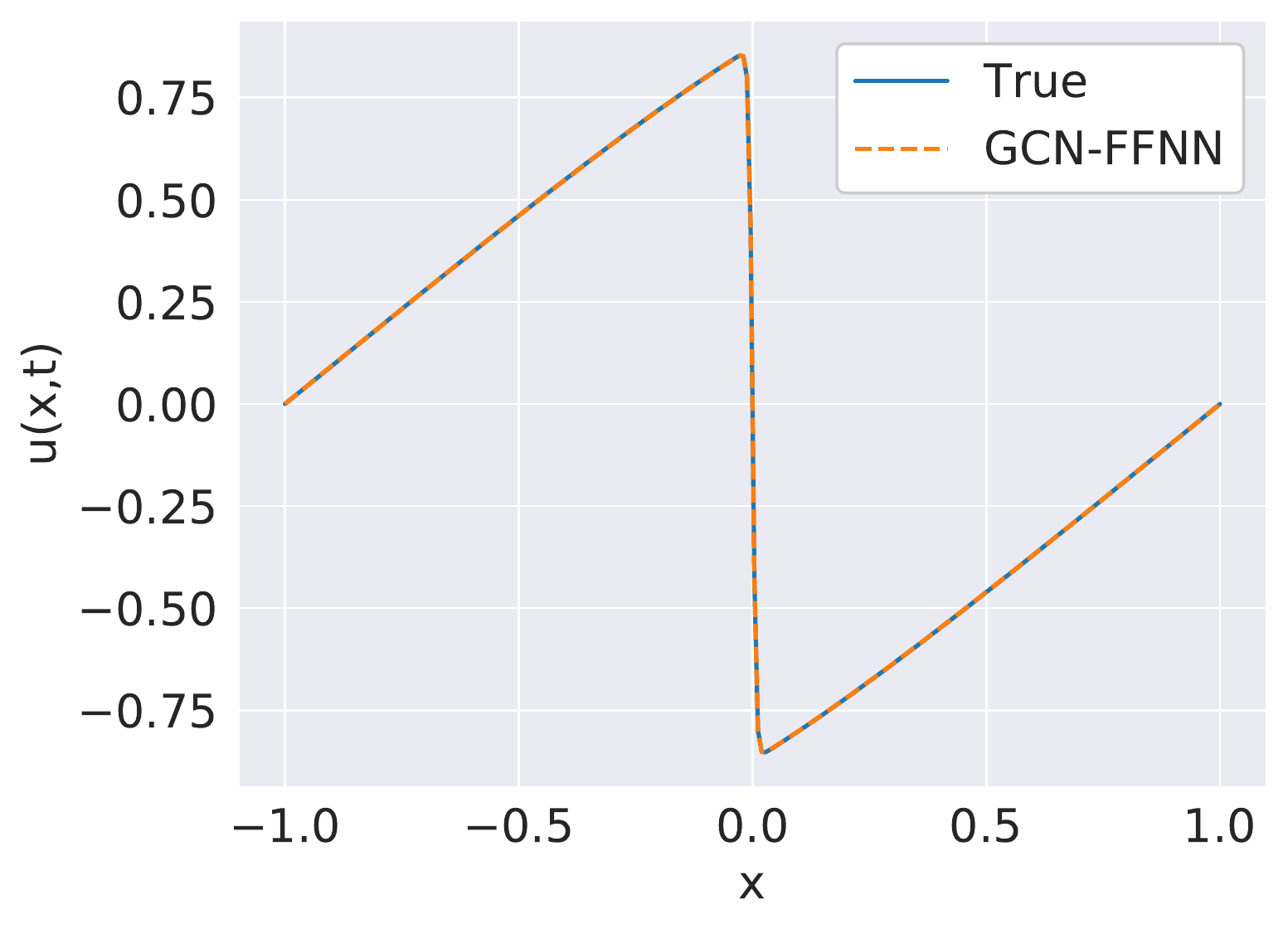}
    \vspace{-.11\textwidth}\caption{}\vspace{.05\textwidth}
    \label{fig:t075_outside_1D_burgers}
  \end{subfigure}\hspace{0.02\textwidth}
  \begin{subfigure}{.3\textwidth}
    \figuretitle{The plots of $u(x,t)$ at $t=0.99$}
    \includegraphics[width=\textwidth]{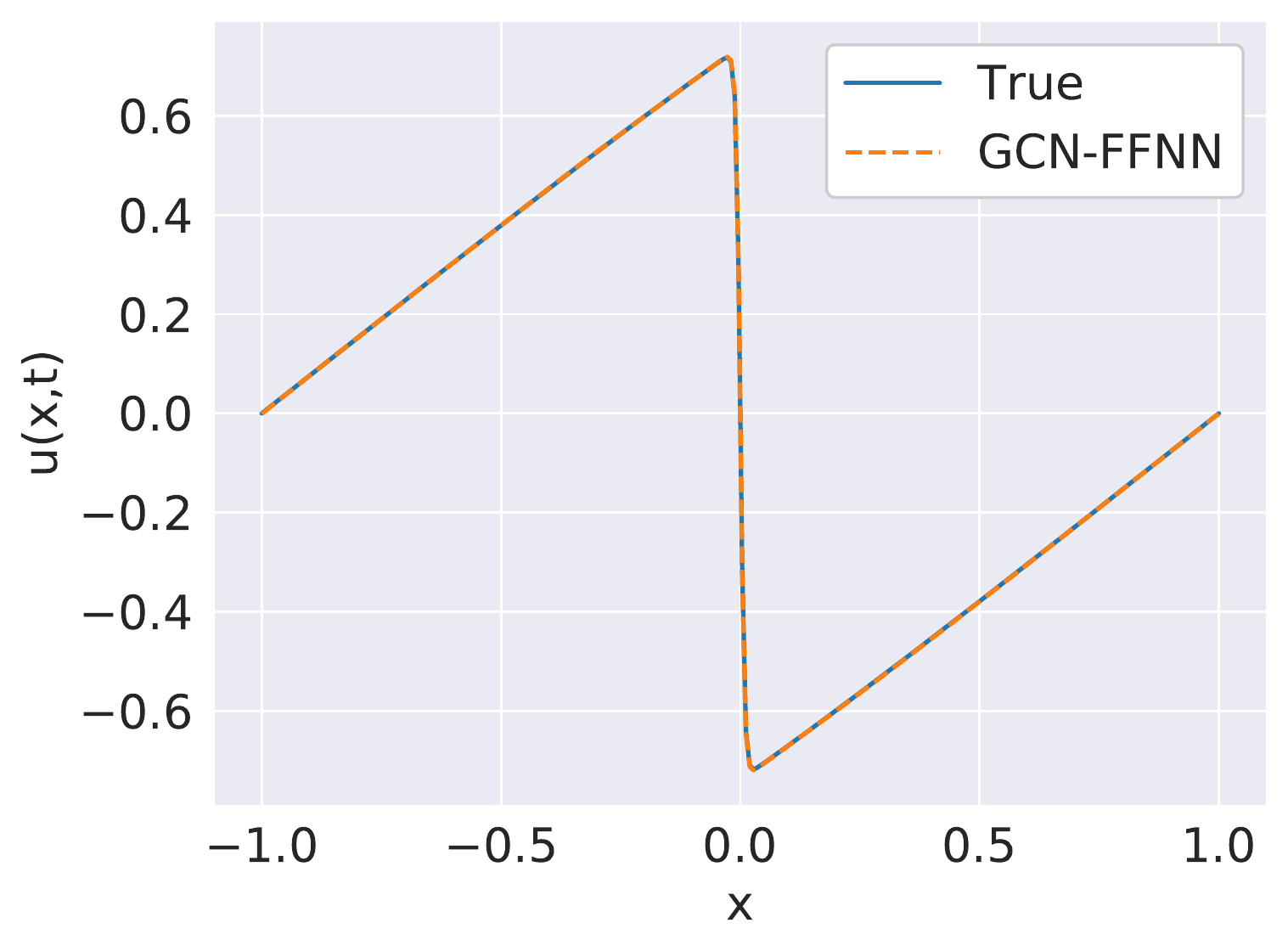}
    \vspace{-.11\textwidth}\caption{}\vspace{.05\textwidth}
    \label{fig:t099_outside_1D_burgers}
  \end{subfigure}
 \begin{subfigure}{.3\textwidth}
    \figuretitle{The plot of $u(x,t)-\hat{u}(x,t)$ at $t=0.50$}
    \includegraphics[width=\textwidth]{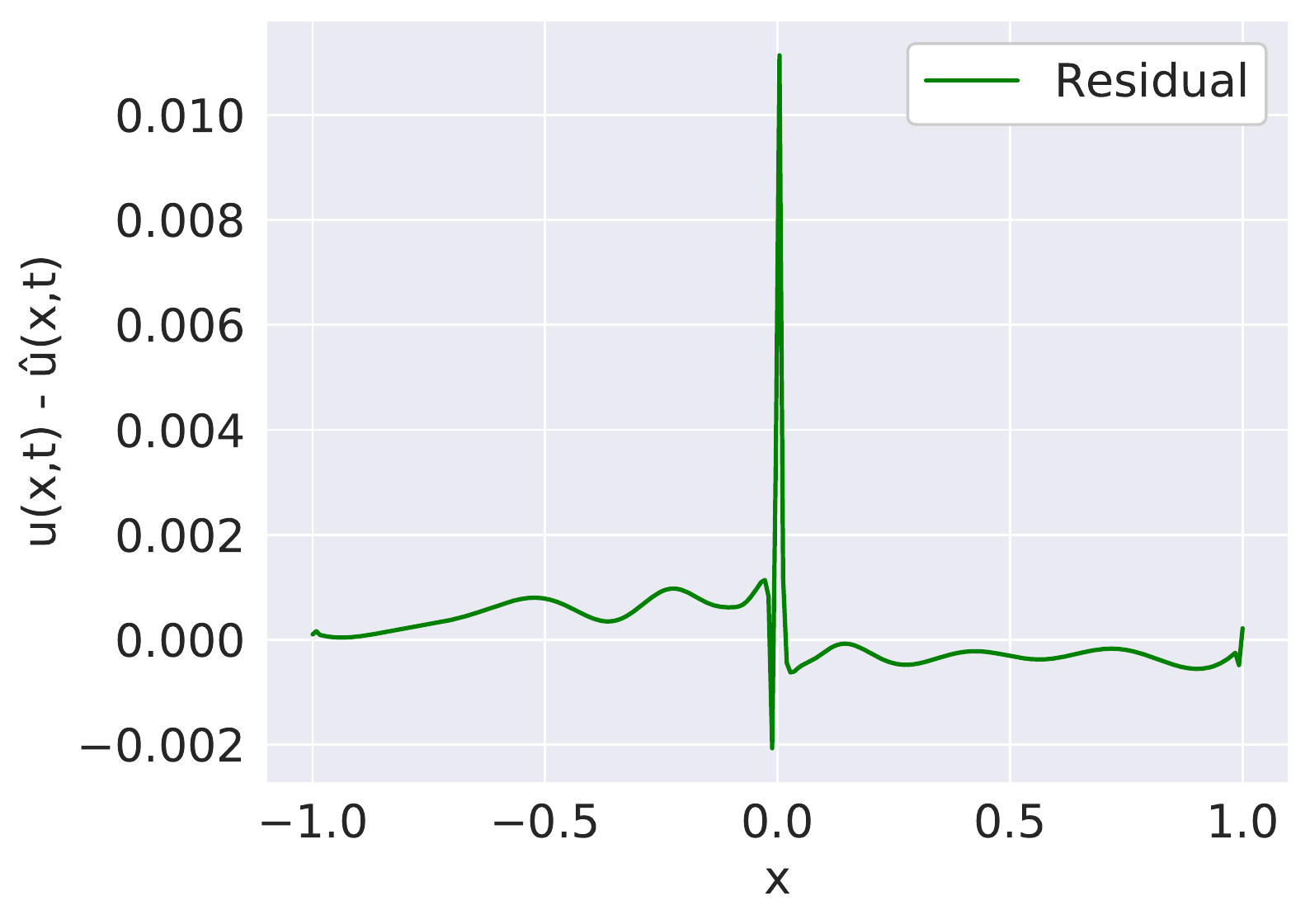}
    \vspace{-.11\textwidth}\caption{}
    \label{fig:r050_outside_1D_burgers}
  \end{subfigure}\hspace{0.02\textwidth}
  \begin{subfigure}{.3\textwidth}
    \figuretitle{The plot of $u(x,t)-\hat{u}(x,t)$ at $t=0.75$}
    \includegraphics[width=\textwidth]{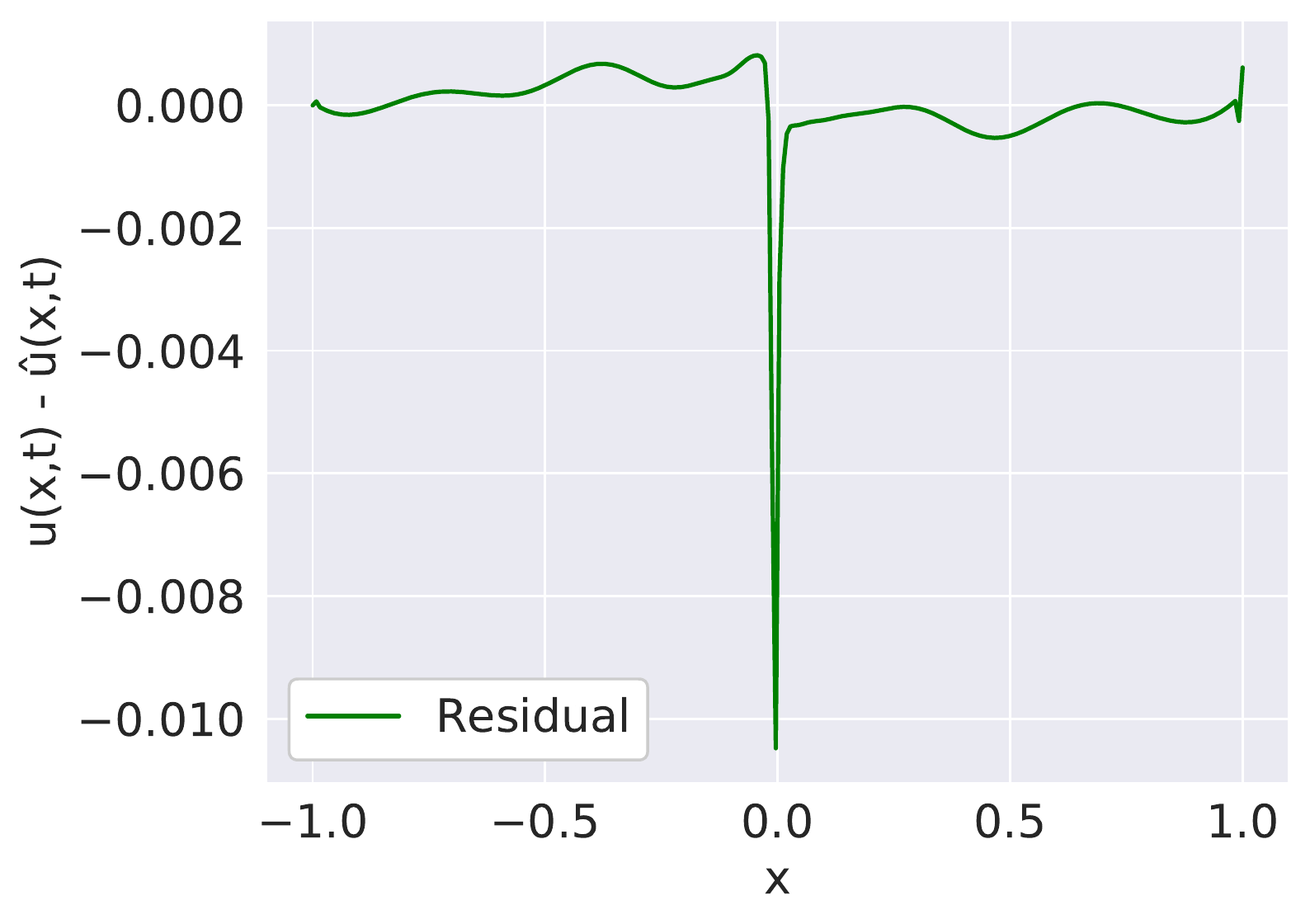}
    \vspace{-.11\textwidth}\caption{}
    \label{fig:r075_outside_1D_burgers}
  \end{subfigure}\hspace{0.02\textwidth}
  \begin{subfigure}{.3\textwidth}
    \figuretitle{The plot of $u(x,t)-\hat{u}(x,t)$ at $t=0.99$}
    \includegraphics[width=\textwidth]{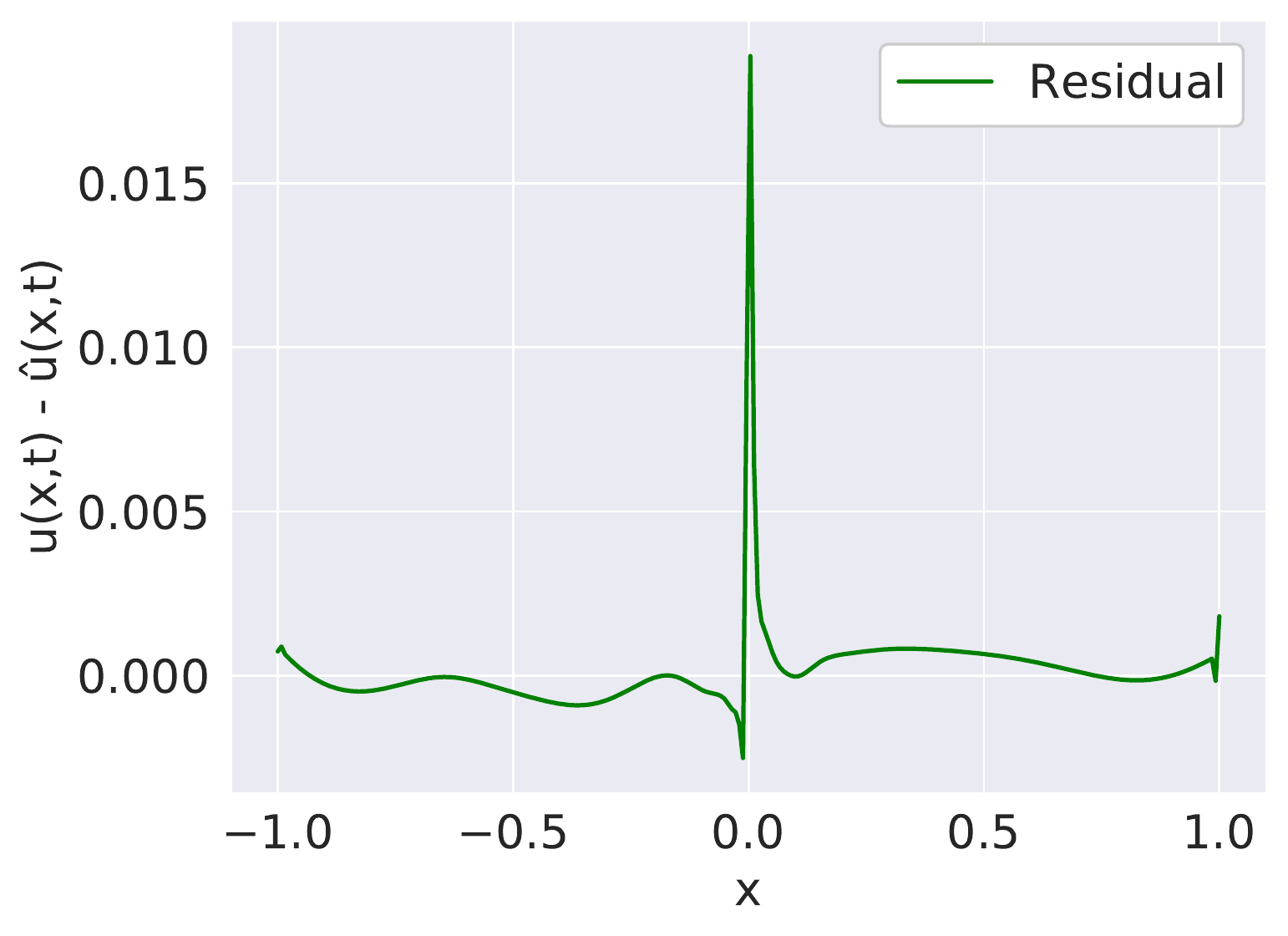}
    \vspace{-.11\textwidth}\caption{}
    \label{fig:r099_outside_1D_burgers}
  \end{subfigure}
    \caption{\textbf{Second scenario:} The plots for the 1D-Burgers equation obtained by GCN-FFNN model corresponding to the second scenario where test nodes are from outside the domain. (a),(b) and (c): The true and approximate solution obtained by GCN-FFNN model. (d),(e) and (f): The obtained residuals $u(x,t)-\hat{u}(x,t)$. The data at $t=0.50$ and $t=0.75$ belong to the training set, while the data at $t=0.99$ is from test set.}
  \label{fig:1d-burgers_plots_outside}
\end{figure*}

Fig. \ref{fig:1d-burgers_plots_inside} corresponds to the first scenario where the test nodes are from inside the domain. In particular, Fig. \ref{fig:1d-burgers_plots_inside} (a), (b) and (c) show the true and approximate solution obtained by GCN-FFNN model for the 1D-Burgers equation at $x=-0.15$, $x=0.15$ and $x=0.94$. Here, the prediction at $x=-0.15$ is from training set, whereas the predictions at $x=0.15$ and $x=0.94$ are from test set. The obtained residuals are shown in Fig. \ref{fig:1d-burgers_plots_inside} (d), (e) and (f).  Fig. \ref{fig:1d-burgers_plots_outside} corresponds to the second scenario where the test nodes are from outside the domain.  Fig. \ref{fig:1d-burgers_plots_outside} (a), (b) and (c) show the true and approximate solution obtained by GCN-FFNN model. Here, the predictions at $t=0.50$ and $t=0.75$ are from the train set, while the predictions at $t=0.99$ are from the test set. Fig. \ref{fig:1d-burgers_plots_outside} (d), (e) and (f) are the residuals at $t=0.50$, $t=0.75$ and $t=0.99$.

\subsection{1D-Schrödinger Equation}\label{Burgers Equation}

\begin{figure*}[hbt!]
\center{}
  \begin{subfigure}{.3\textwidth}
    \figuretitle{The plots of $u(x,t)$ at $x=-1.17$}
    \includegraphics[width=\textwidth]{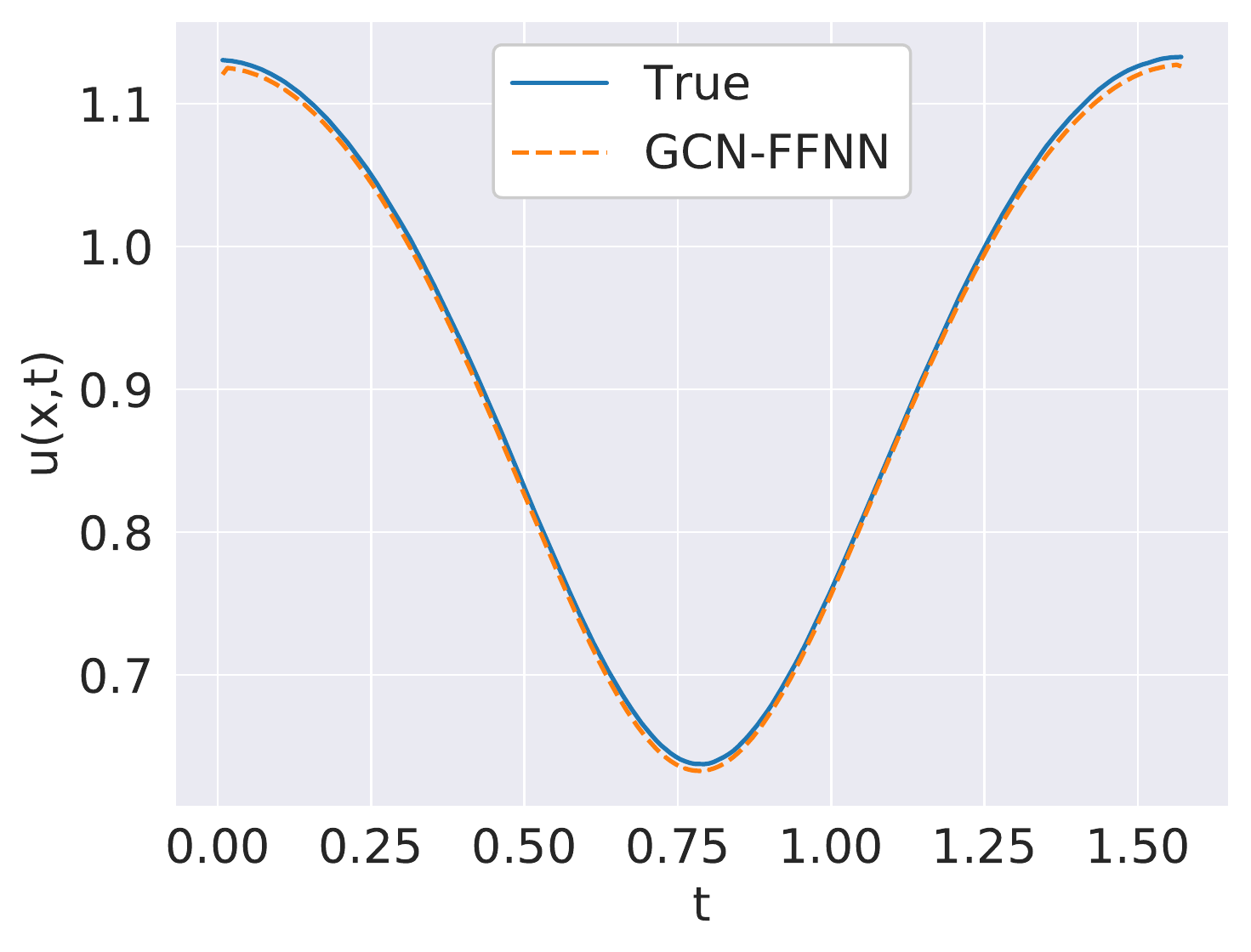}
    \vspace{-.11\textwidth}\caption{}\vspace{.05\textwidth}
    \label{fig:x1_inside_1D_schrödinger}
  \end{subfigure}\hspace{0.02\textwidth}
  \begin{subfigure}{.3\textwidth}
    \figuretitle{The plots of $u(x,t)$ at $x=0$}
    \includegraphics[width=\textwidth]{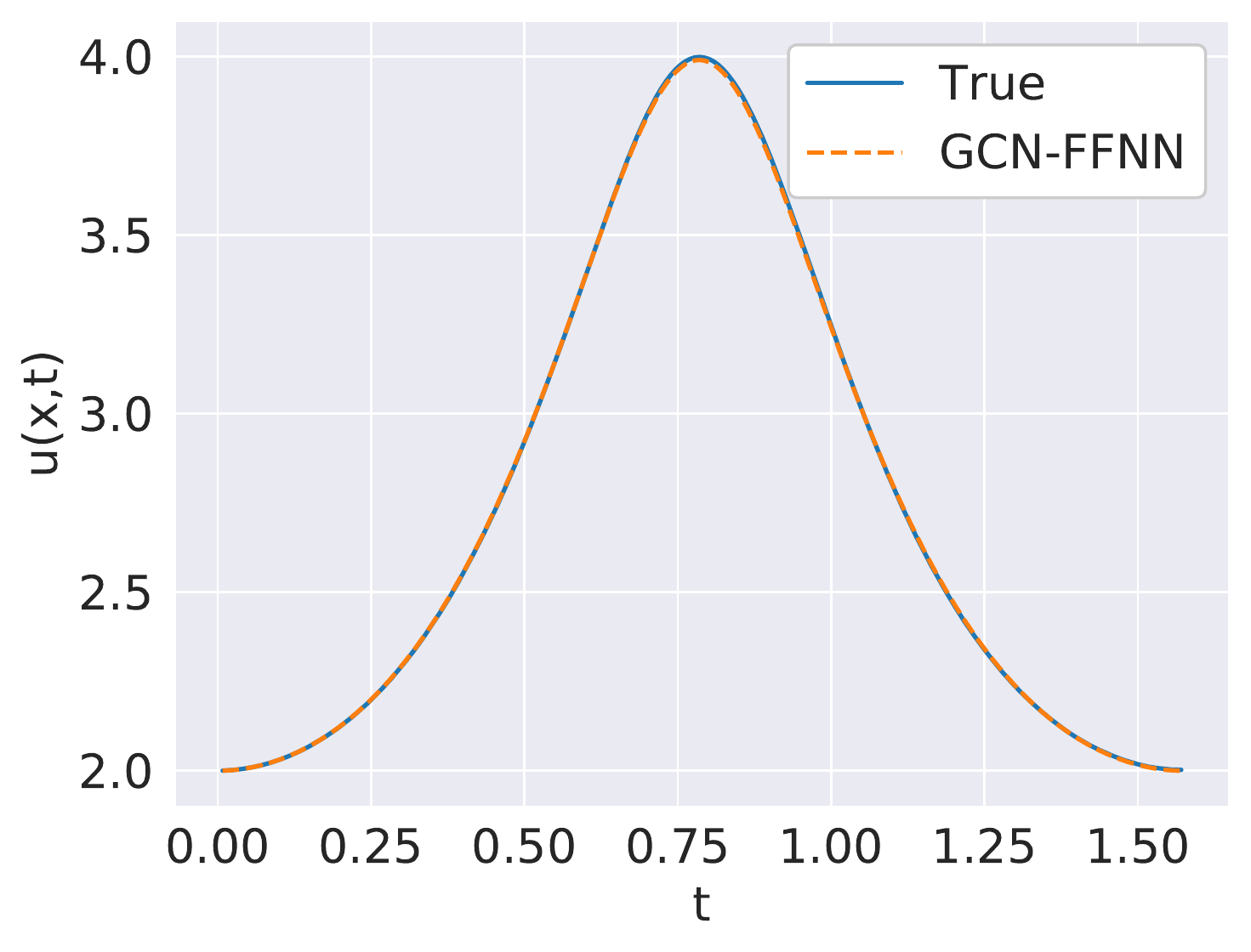}
    \vspace{-.11\textwidth}\caption{}\vspace{.05\textwidth}
    \label{fig:x2_inside_1D_schrödinger}
  \end{subfigure}\hspace{0.02\textwidth}
  \begin{subfigure}{.3\textwidth}
    \figuretitle{The plots of $u(x,t)$ at $x=1.56$}
    \includegraphics[width=\textwidth]{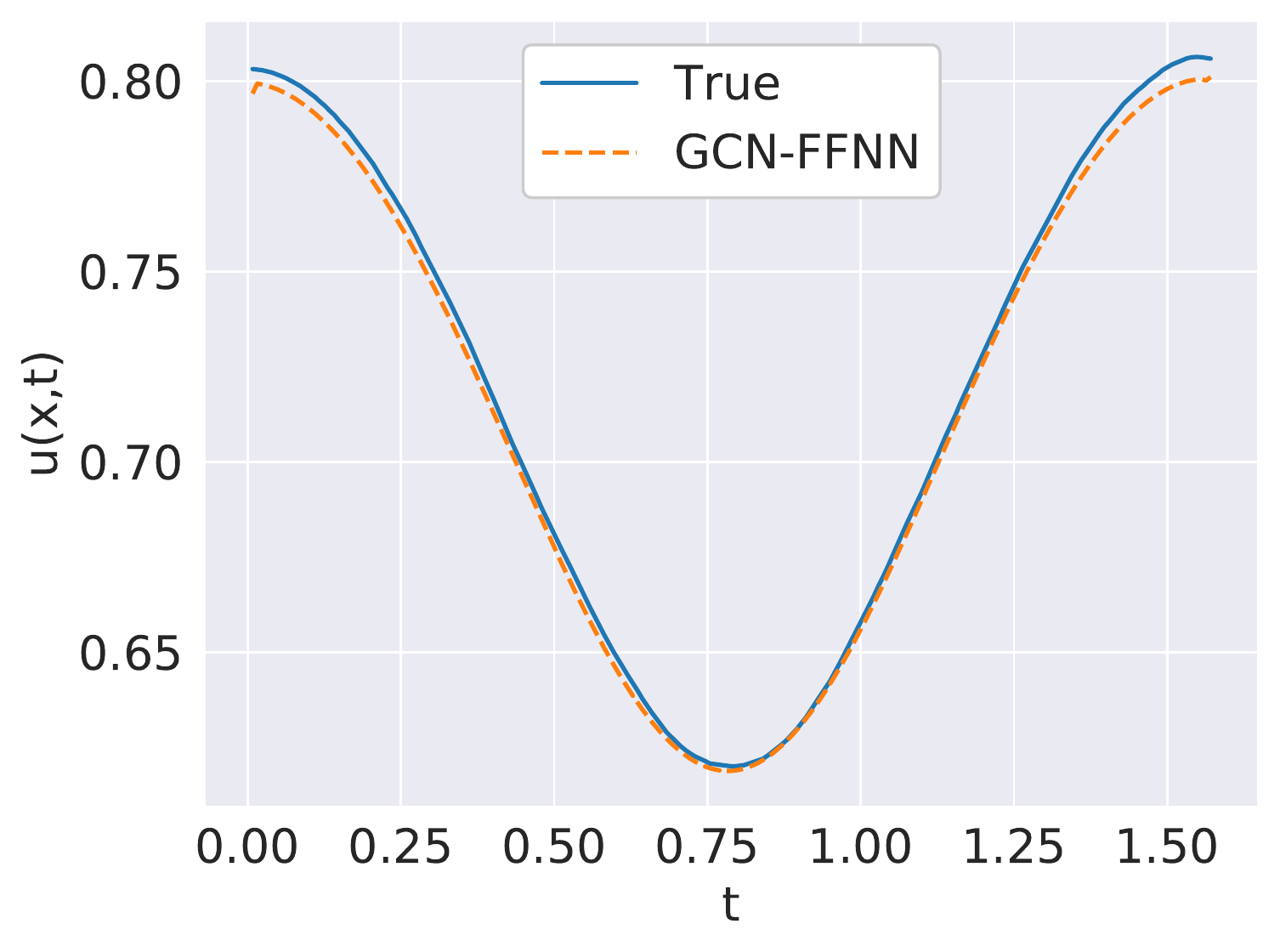}
    \vspace{-.11\textwidth}\caption{}\vspace{.05\textwidth}
    \label{fig:x3_inside_1D_schrödinger}
  \end{subfigure}
   \begin{subfigure}{.3\textwidth}
    \figuretitle{The plot of $u(x,t)-\hat{u}(x,t)$ at $x=-1.17$}
    \includegraphics[width=\textwidth]{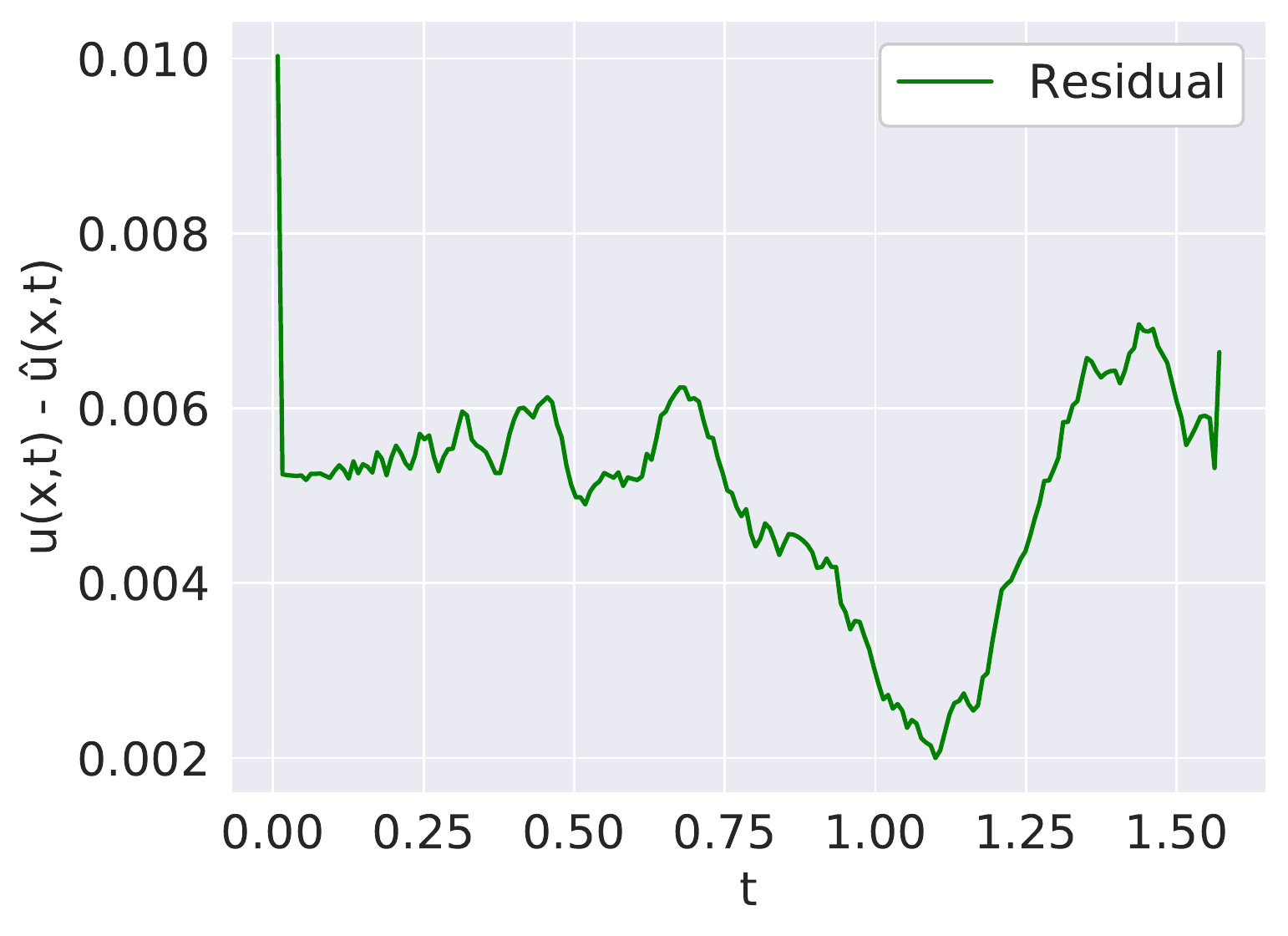}
    \vspace{-.11\textwidth}\caption{}
    \label{fig:r1_inside_1D_schrödinger}
  \end{subfigure}\hspace{0.02\textwidth}
  \begin{subfigure}{.3\textwidth}
    \figuretitle{The plot of $u(x,t)-\hat{u}(x,t)$ at $x=0$}
    \includegraphics[width=\textwidth]{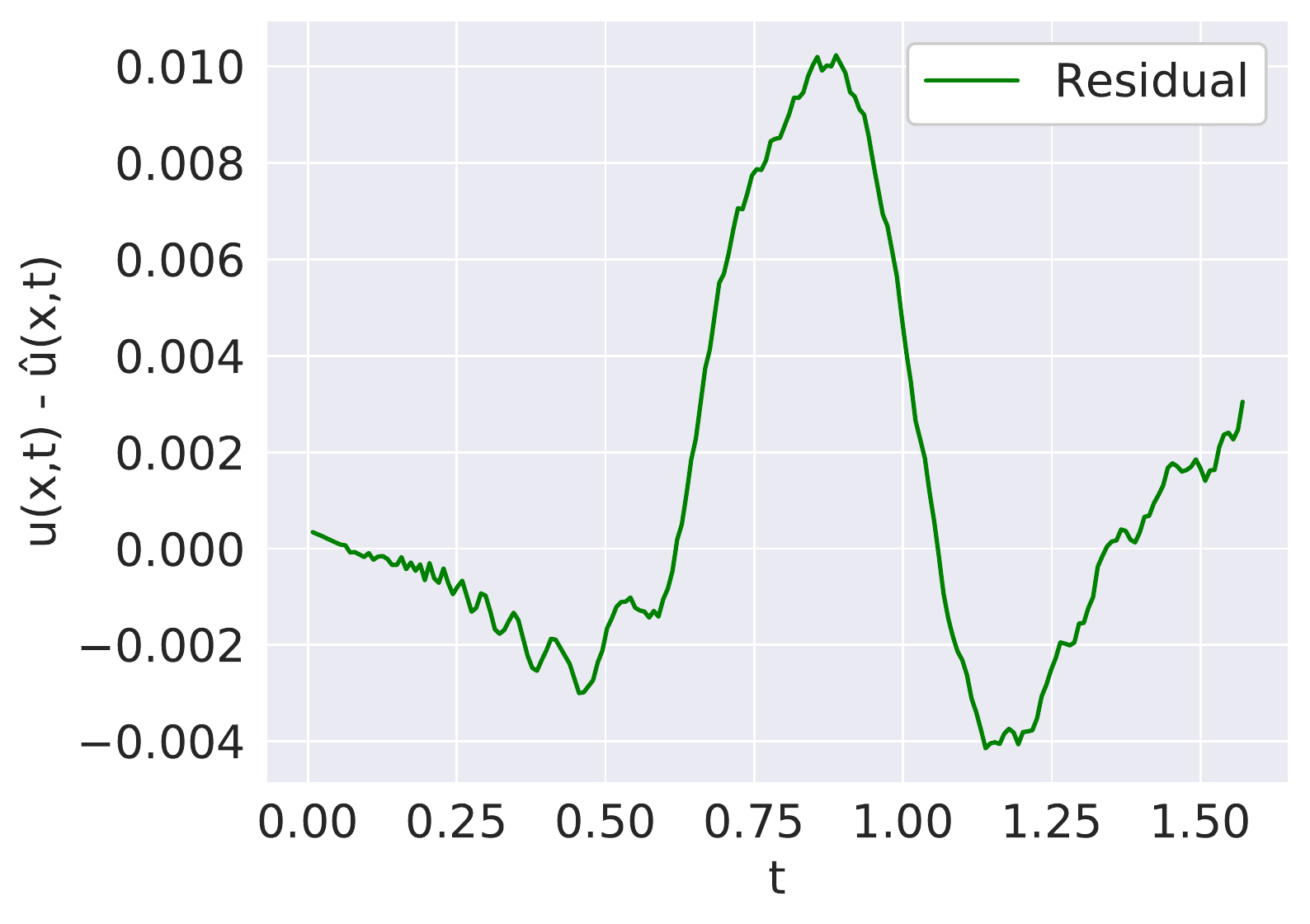}
    \vspace{-.11\textwidth}\caption{}
    \label{fig:r2_inside_1D_schrödinger}
  \end{subfigure}\hspace{0.02\textwidth}
  \begin{subfigure}{.3\textwidth}
    \figuretitle{The plot of $u(x,t)-\hat{u}(x,t)$ at $x=1.56$}
    \includegraphics[width=\textwidth]{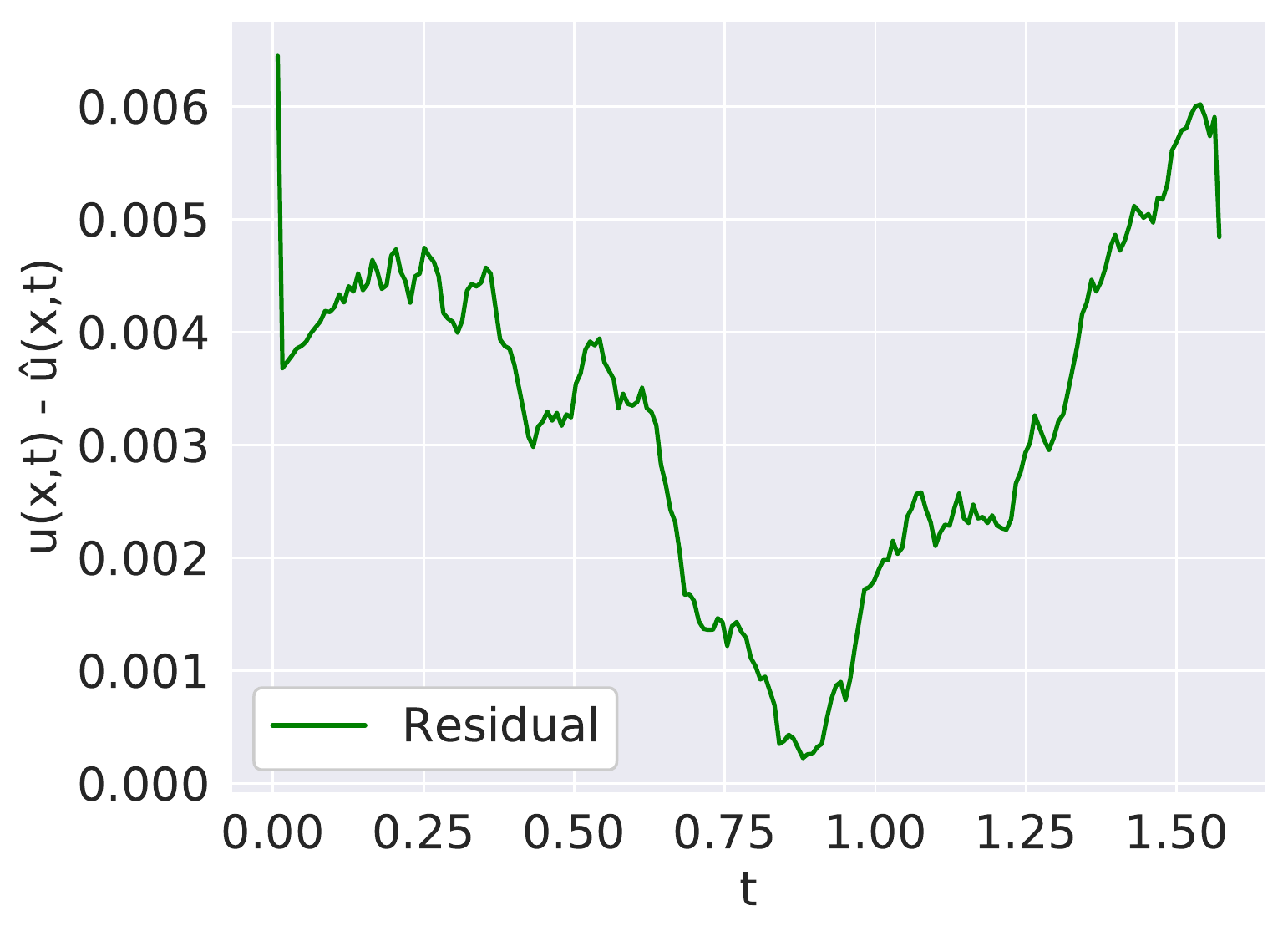}
    \vspace{-.11\textwidth}\caption{}
    \label{fig:r3_inside_1D_schrödinger}
  \end{subfigure}
  \caption{\textbf{First scenario:} The plots for the 1D-Schrödinger equation obtained by GCN-FFNN model corresponding to the first scenario where test nodes are from inside the domain. (a),(b) and (c): The true and approximate solution obtained by GCN-FFNN model. (d),(e) and (f): The obtained residuals $u(x,t)-\hat{u}(x,t)$.  The data at $x=-1.17$ belongs to the training set, while the data at $x=0$ and $x=1.56$ are from test set.}
    \label{fig:1d-schrödinger_plots_inside}
\end{figure*}

\begin{figure*}[hbt!]
\center{}
  \begin{subfigure}{.3\textwidth}
    \figuretitle{The plots of $u(x,t)$ at $t=0.11$}
    \includegraphics[width=\textwidth]{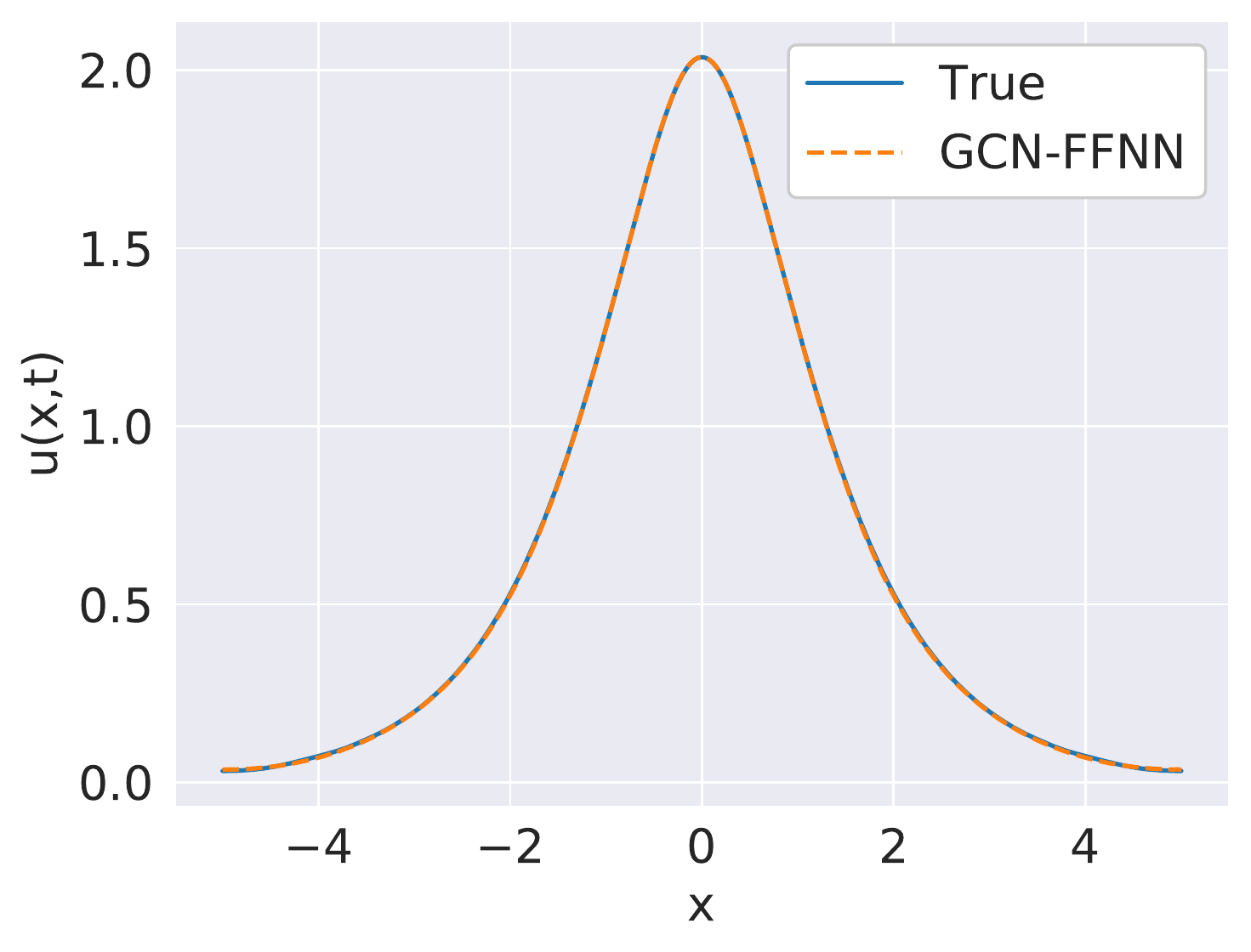}
    \vspace{-.11\textwidth}\caption{}\vspace{.05\textwidth}
    \label{fig:t011_outside_1D_schrödinger}
  \end{subfigure}\hspace{0.02\textwidth}
  \begin{subfigure}{.3\textwidth}
    \figuretitle{The plots of $u(x,t)$ at $t=0.80$}
    \includegraphics[width=\textwidth]{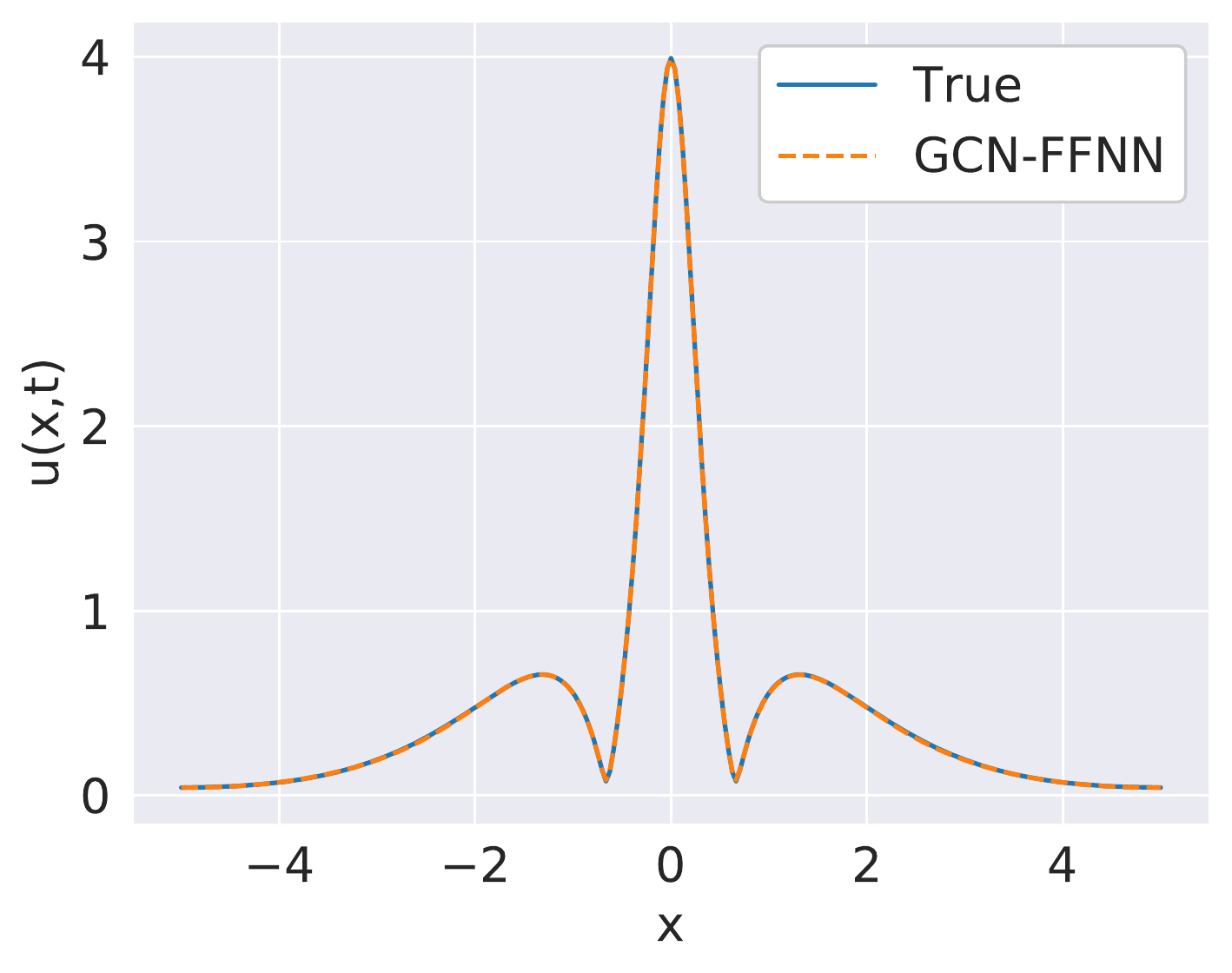}
    \vspace{-.11\textwidth}\caption{}\vspace{.05\textwidth}
    \label{fig:t080_outside_1D_schrödinger}
  \end{subfigure}\hspace{0.02\textwidth}
  \begin{subfigure}{.3\textwidth}
    \figuretitle{The plots of $u(x,t)$ at $t=1.57$}
    \includegraphics[width=\textwidth]{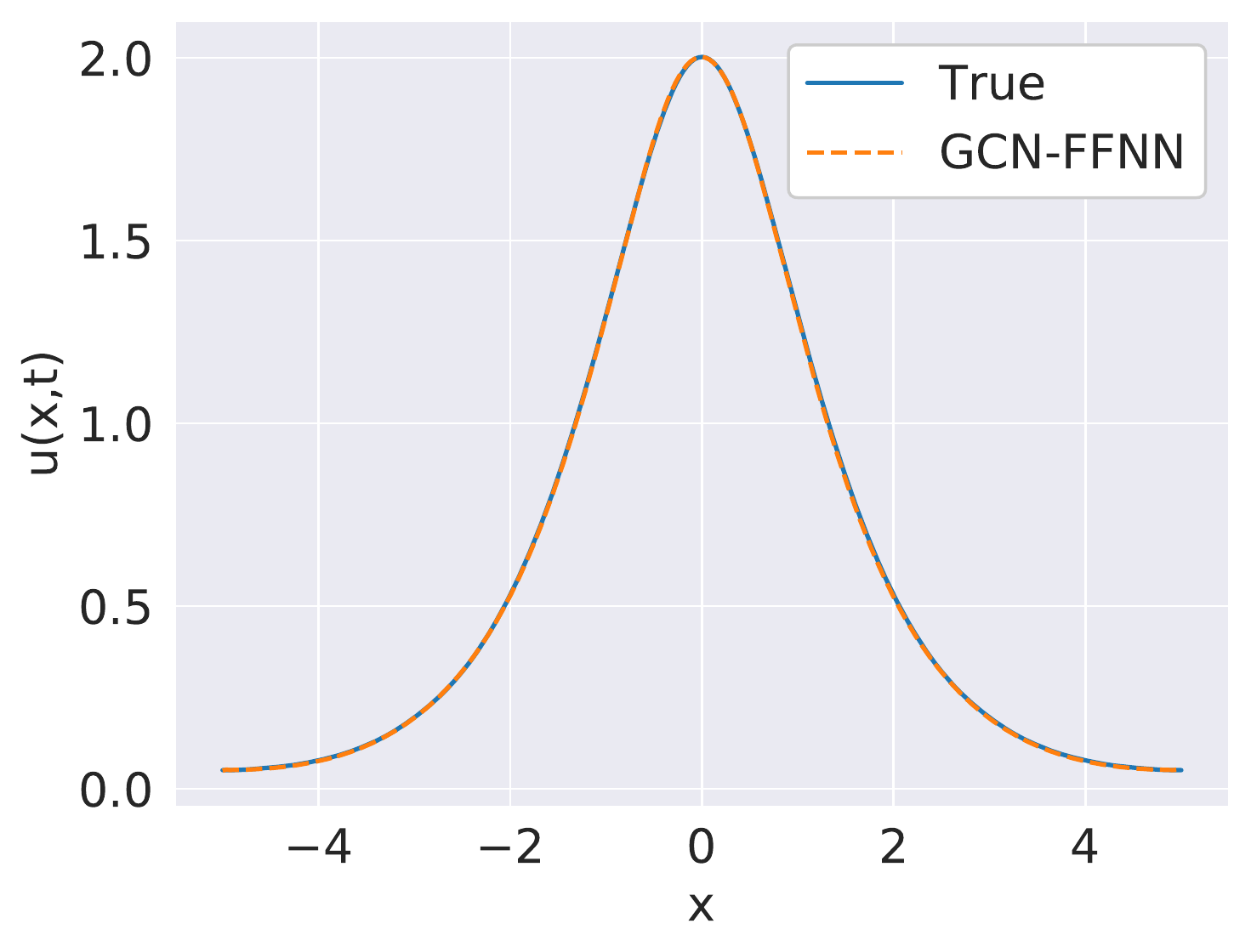}
    \vspace{-.11\textwidth}\caption{}\vspace{.05\textwidth}
    \label{fig:t157_outside_1D_schrödinger}
  \end{subfigure}
  \begin{subfigure}{.3\textwidth}
    \figuretitle{The plot of $u(x,t)-\hat{u}(x,t)$ at $t=0.11$}
    \includegraphics[width=\textwidth]{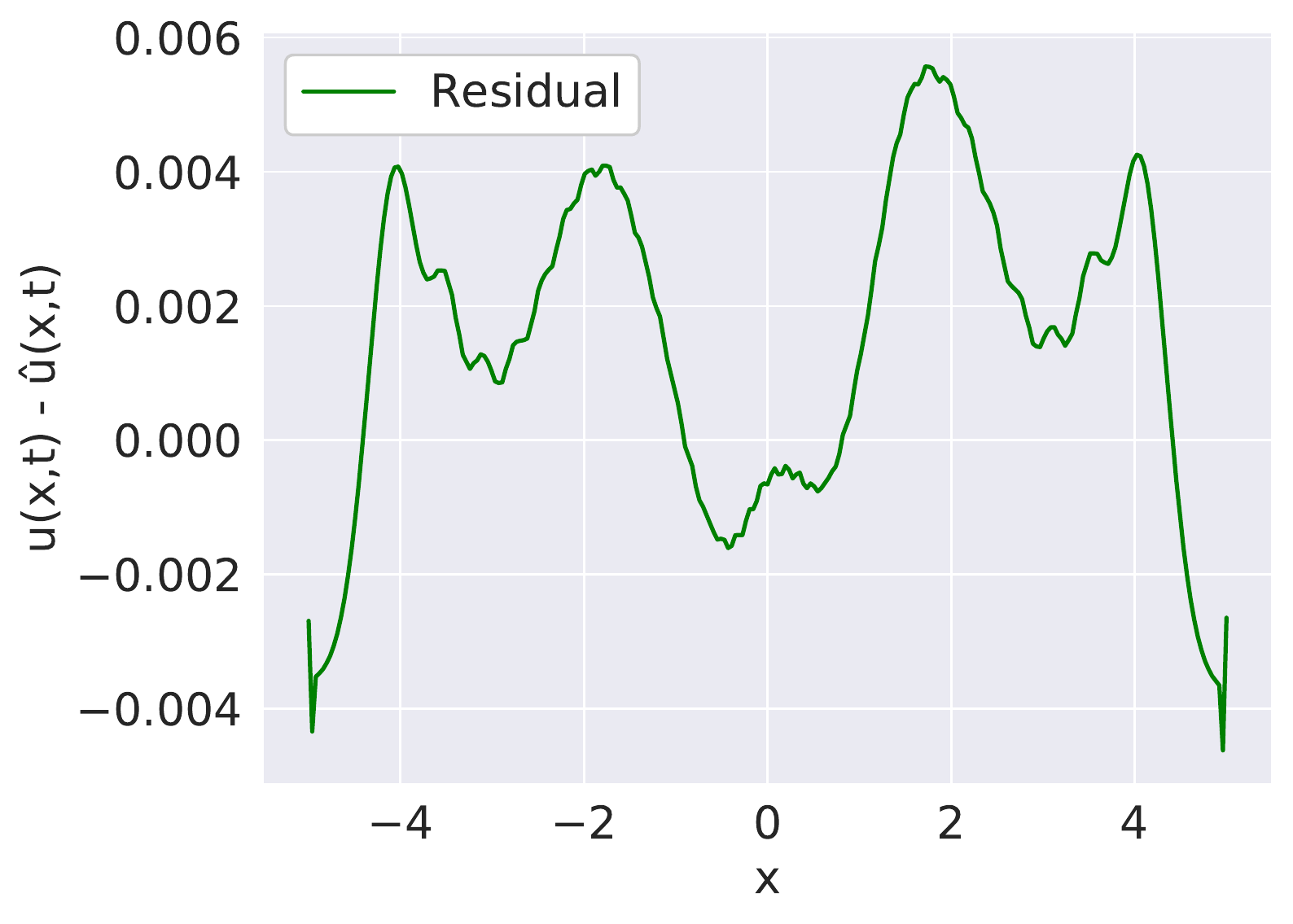}
    \vspace{-.11\textwidth}\caption{}
    \label{fig:r011_outside_1D_schrödinger}
  \end{subfigure}\hspace{0.02\textwidth}
  \begin{subfigure}{.3\textwidth}
    \figuretitle{The plot of $u(x,t)-\hat{u}(x,t)$ at $t=0.80$}
    \includegraphics[width=\textwidth]{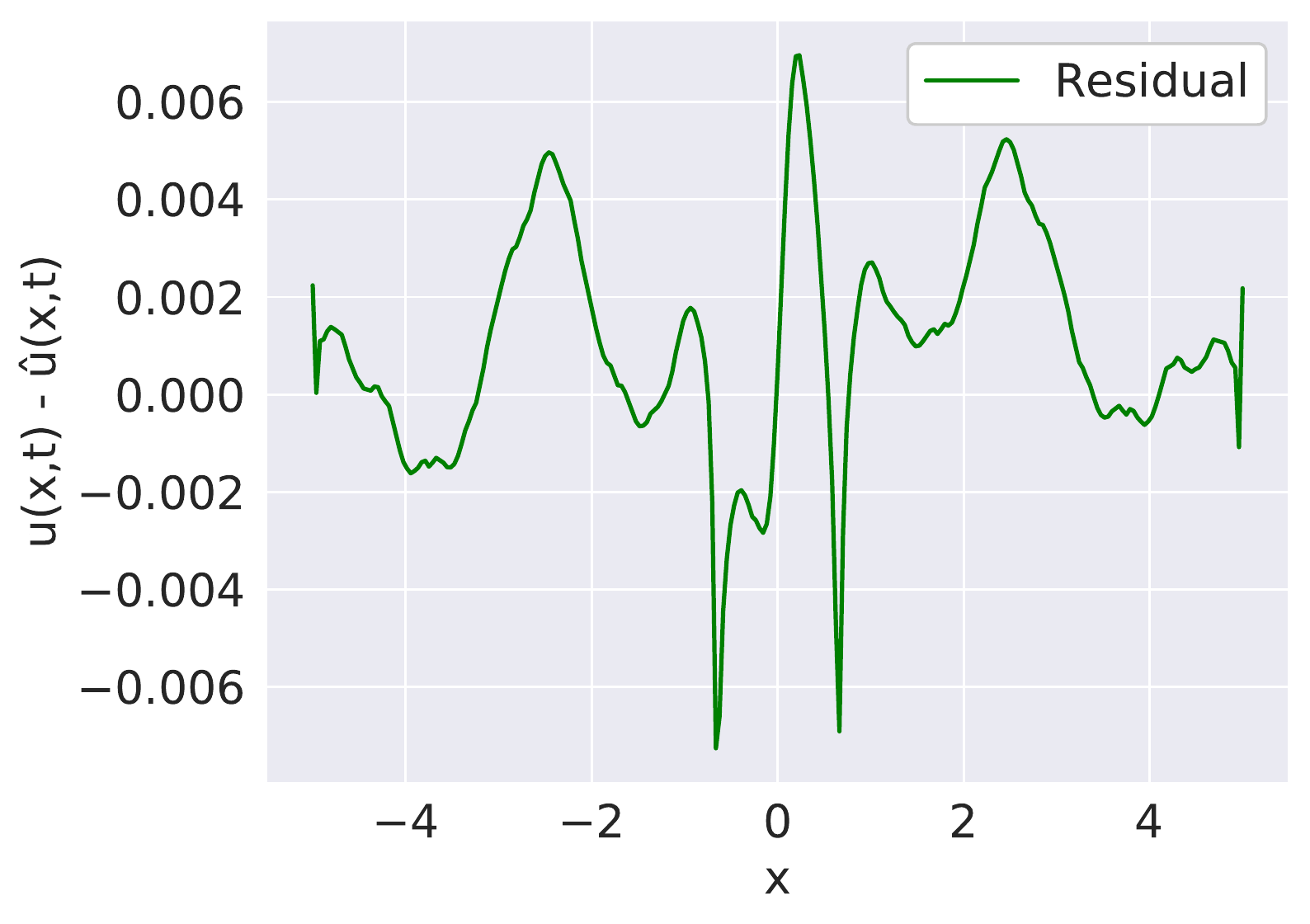}
    \vspace{-.11\textwidth}\caption{}
    \label{fig:r080_outside_1D_schrödinger}
  \end{subfigure}\hspace{0.02\textwidth}
  \begin{subfigure}{.3\textwidth}
    \figuretitle{The plot of $u(x,t)-\hat{u}(x,t)$ at $t=1.57$}
    \includegraphics[width=\textwidth]{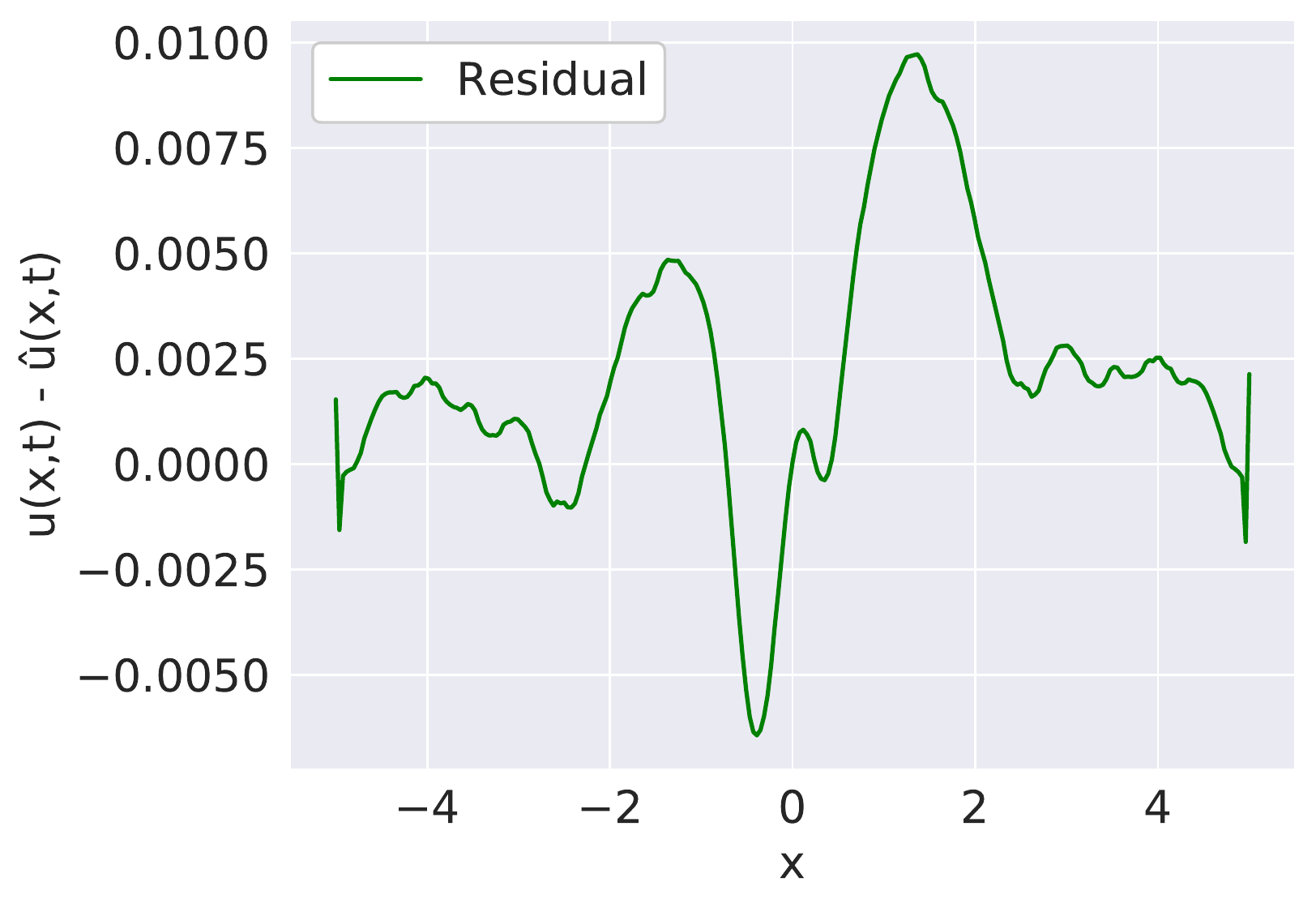}
    \vspace{-.11\textwidth}\caption{}
    \label{fig:r157_outside_1D_schrödinger}
  \end{subfigure}
  \caption{\textbf{Second scenario:} The plots for the 1D-Schrödinger equation obtained by GCN-FFNN model corresponding to the second scenario where test nodes are from outside the domain. (a),(b) and (c): The true and approximate solution obtained by GCN-FFNN model. (d),(e) and (f): The obtained residuals $u(x,t)-\hat{u}(x,t)$. The data at $t=0.11$ and $t=0.80$ belong to the training set, while the data at $t=1.57$ is from test set.}
    \label{fig:1d-schrödinger_plots_outside}
\end{figure*}

The 1D-Schrödinger equation with boundary and initial conditions is described in Eq. (\ref{eq:1d-schrödinger-1}) \cite{li2021deep}.

\begin{equation} \label{eq:1d-schrödinger-1}
\begin{cases}

i\psi_t + 0.5\psi_{xx} + |\psi|^2 \psi = 0,\; x \in [-5,5],\; t\in [0,\frac{\pi}{2}], \\
\psi(x,0)=2sech(x) \\
\psi(-5,t)=\psi(5,t) \\
\psi_x(-5,t)=\psi_x(5,t) \\

\end{cases}
\end{equation}

The differential operator is defined as $f[u]:=f=i\psi_t + 0.5\psi_{xx} + |\psi|^2 \psi$. Let $u$ and $v$ denote the real and imaginary components of $\psi$, then the 1D-Schrödinger equation can be rewritten as follows \cite{li2021deep}: 

\begin{equation} \label{eq:1d-schrödinger-3}
\begin{cases}
u_t = -0.5v_{xx}-(u^2+v^2)v, \\
v_t = 0.5u_{xx}+(u^2+v^2)u. \\

\end{cases}
\end{equation}

In the first scenario, the domain of the 1D-Schrödinger equation $(x,t) \in [-5,5]\times [0,\frac{\pi}{2}]$ is divided into $N=257\times201$ nodes, which are evenly spaced for each dimension. Following the previously mentioned  approaches for creating the test nodes, $10\%$ of $N$ nodes are randomly selected to form the inside domain test nodes. In the second scenario, the nodes in the ranges $(x,t) \in [-5,5]\times [0,0.9\frac{\pi}{2}]$ are used for training and the outside domain test nodes are selected from $0.9\frac{\pi}{2}< t\leq \frac{\pi}{2}$. 
The obtained results for the 1D-Schrödinger equation on both inside and outside domain test nodes are shown in Table \ref{tab:loss-inside}. FFNN outperforms the other models for inside domain test nodes. For the outside domain test nodes, the GCN-FFNN model shows the least MSE error compared to the other models, while its infinity norm error remains higher compared to FFNN. 

Fig. \ref{fig:1d-schrödinger_plots_inside} corresponds to the first scenario where the test nodes are from inside the domain.
In Fig. \ref{fig:1d-schrödinger_plots_inside} (a), (b) and (c) the true solution and the obtained results of GCN-FFNN model at $x=-1.17$, $x=0$ and $x=1.56$ are shown, respectively. Here, the prediction at $x=-1.17$ is from training set, whereas the predictions at $x=0$ and $x=1.56$ are from test set. The obtained residuals are shown in Fig. \ref{fig:1d-schrödinger_plots_inside} (d), (e) and (f). Fig. \ref{fig:1d-schrödinger_plots_outside} corresponds to the second scenario where the the test nodes are from outside the domain. The true and approximate solutions at $t=0.11$ $t=0.80$ and $t=1.57$ are shown Fig. \ref{fig:1d-schrödinger_plots_outside} (a), (b) and (c), respectively. Here, the predictions at $t=0.11$ and $t=0.80$ are from the training set, while the predictions at $t=1.57$ are from the outside domain test nodes. The obtained residuals are shown in  Fig. \ref{fig:1d-schrödinger_plots_outside} (d), (e) and (f).

\subsection{2D-Burgers Equation}\label{Burgers Equation}

The 2D-Burgers equation with boundary and initial conditions is given in Eq. (\ref{eq:2d-burgers-1}) \cite{li2021deep}.

\begin{equation} \label{eq:2d-burgers-1}
\begin{cases}
u_t+u(u_x+u_y)-0.1(u_{xx} + u_{yy})=0,\; (x,y) \in [0,1],\; t\in [0,3], \\

u(x,y,0)= 1/(1+\exp{[(x+y-t)/0.2]}), \\

\end{cases}
\end{equation}
Here, the differential operator is defined as $f[u]:=
f=u_t+u(u_x+u_y)-0.1(u_{xx}+u_{yy}).$

The domain of the 2D-Burgers equation, i.e. $(x,y,t) \in [0,1]\times[0,1]\times[0,3]$, is divided into $N=26\times26\times31$ nodes. In the first scenario, $10\%$ of $N$ nodes are randomly selected to form the inside domain test nodes and the remaining nodes are used for training the models. In the second scenario, the nodes in the ranges $(x,y,t) \in [0,1]\times[0,1]\times[0,2.7]$ are used for training and the outside domain test nodes are selected from $2.7 < t\leq 3$. From Table \ref{tab:loss-inside}, one can observe that for the 2D-Burgers equation, the GCN-FFNN model outperforms the other models on inside domain test nodes using both MSE and infinity norms. For outside domain test nodes, GCN outperforms the other models in terms of MSE metric, while the GCN-FFNN model achieved the least infinity norm compared to the other models.

\begin{figure*}[hbt!]
\center{}
  \begin{subfigure}{.35\textwidth}
    \figuretitle{The plots of $u(x,y,t)$ at $t=1$}
    \includegraphics[width=\textwidth]{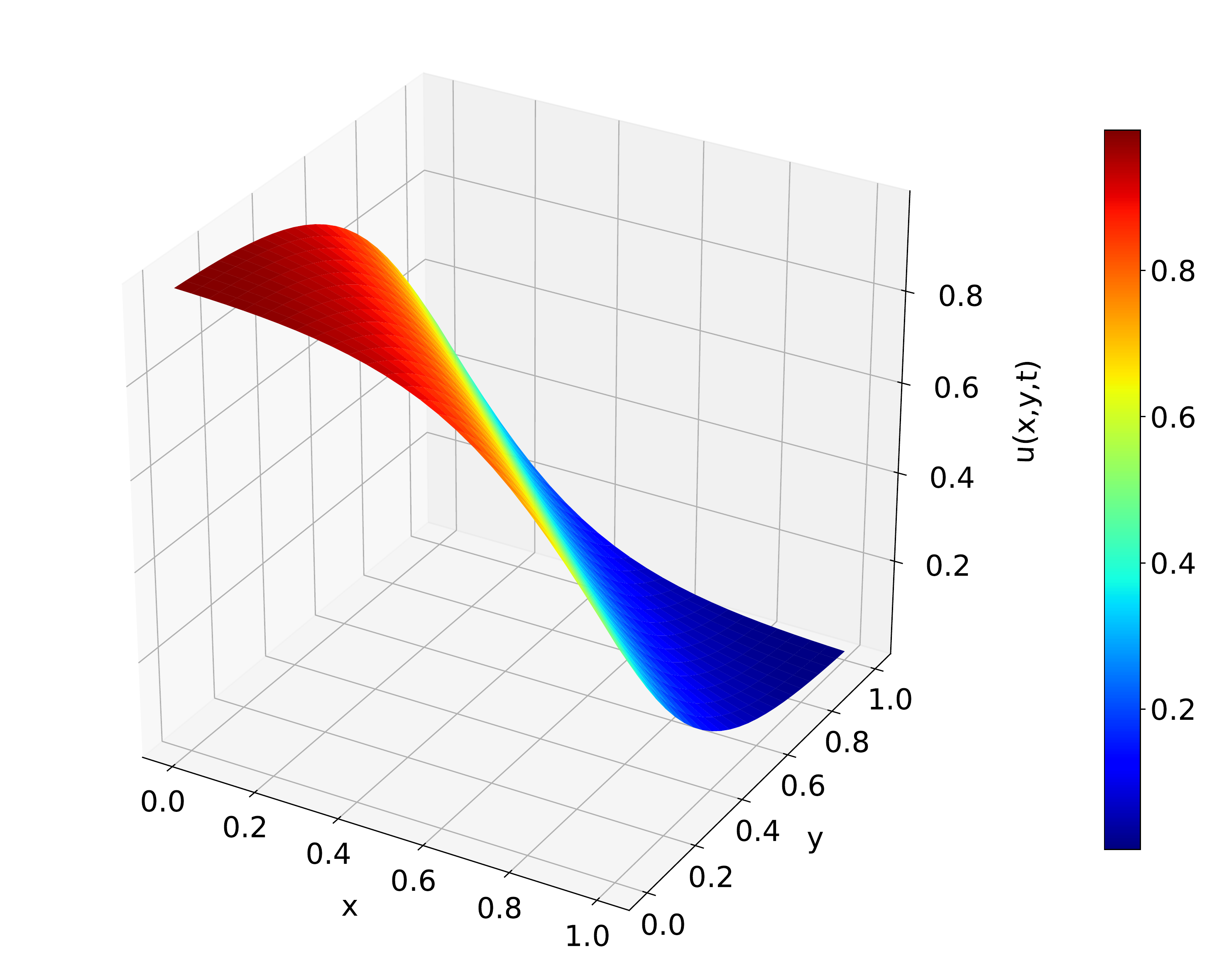}
    \vspace{-.11\textwidth}\caption{}
  \end{subfigure}\hspace{0.02\textwidth}
  \begin{subfigure}{.35\textwidth}
    \figuretitle{The plots of $u(x,y,t)$ at $t=3$}
    \includegraphics[width=\textwidth]{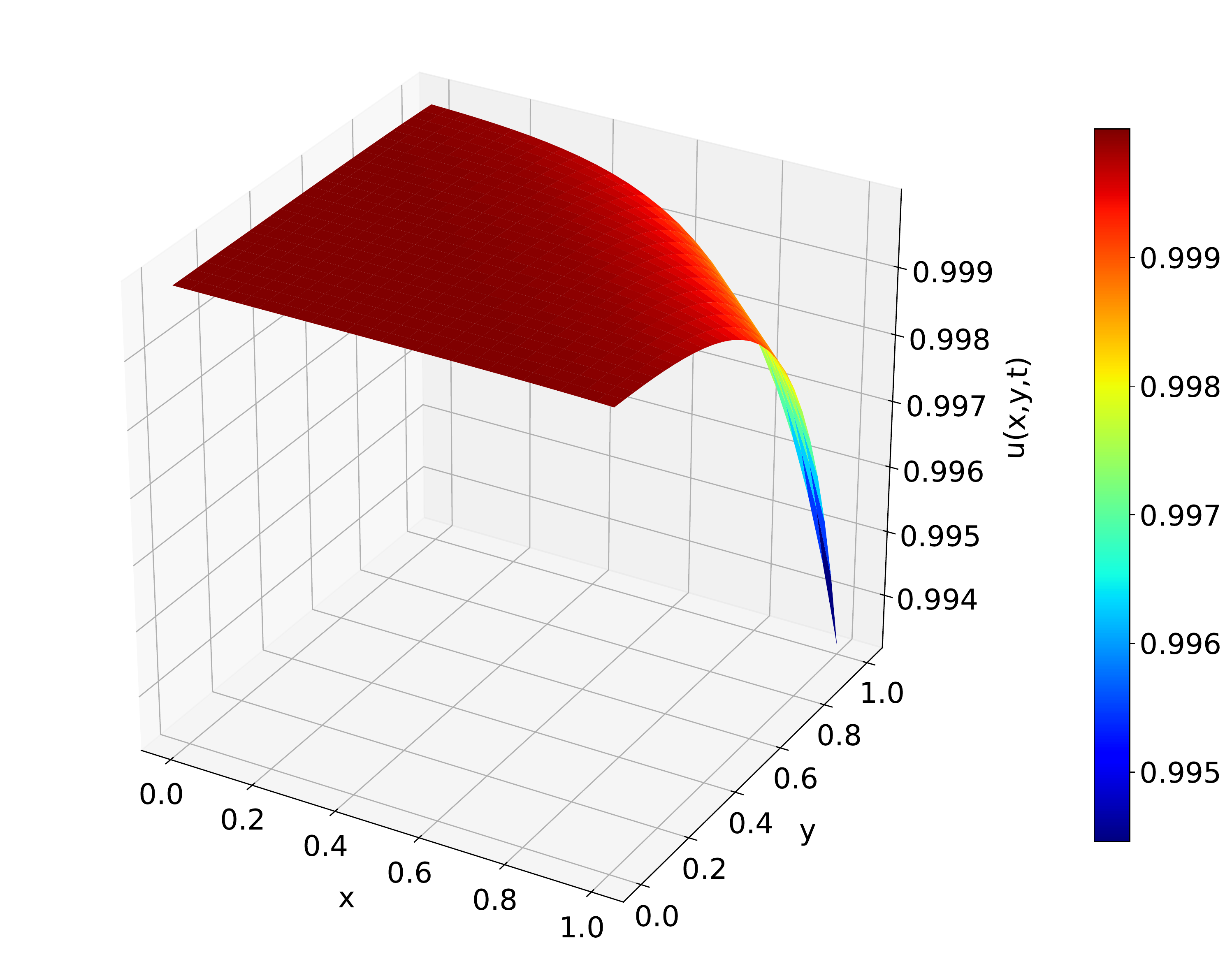}
    \vspace{-.11\textwidth}\caption{}
  \end{subfigure}\hspace{0.02\textwidth}
  \begin{subfigure}{.35\textwidth}
    \figuretitle{The plot of $u(x,y,t)-\hat{u}(x,y,t)$ at $t=1$}
    \includegraphics[width=\textwidth]{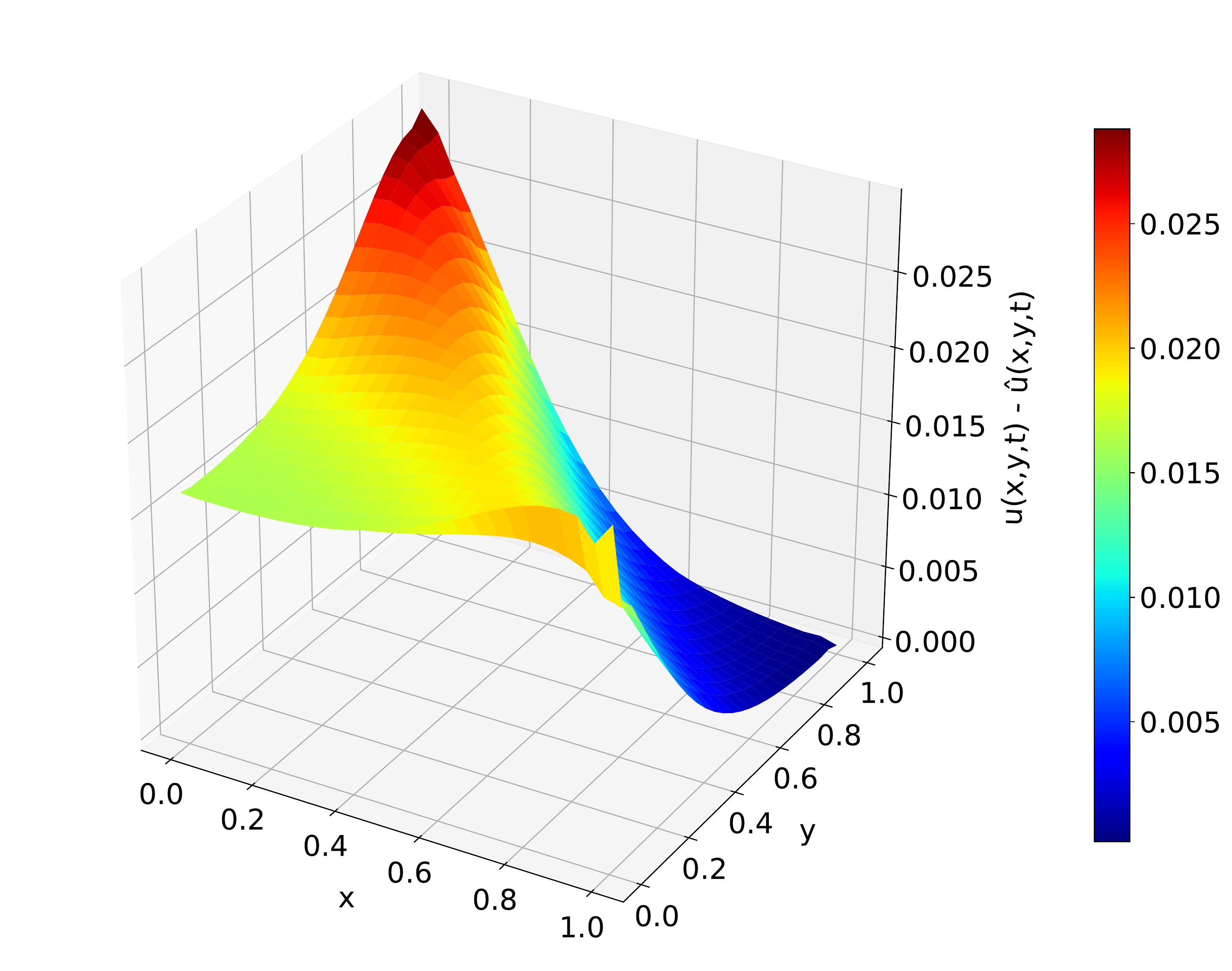}
    \vspace{-.11\textwidth}\caption{}
  \end{subfigure}\hspace{0.02\textwidth}
  \begin{subfigure}{.35\textwidth}
    \figuretitle{The plot of $u(x,y,t)-\hat{u}(x,y,t)$ at $t=3$}
    \includegraphics[width=\textwidth]{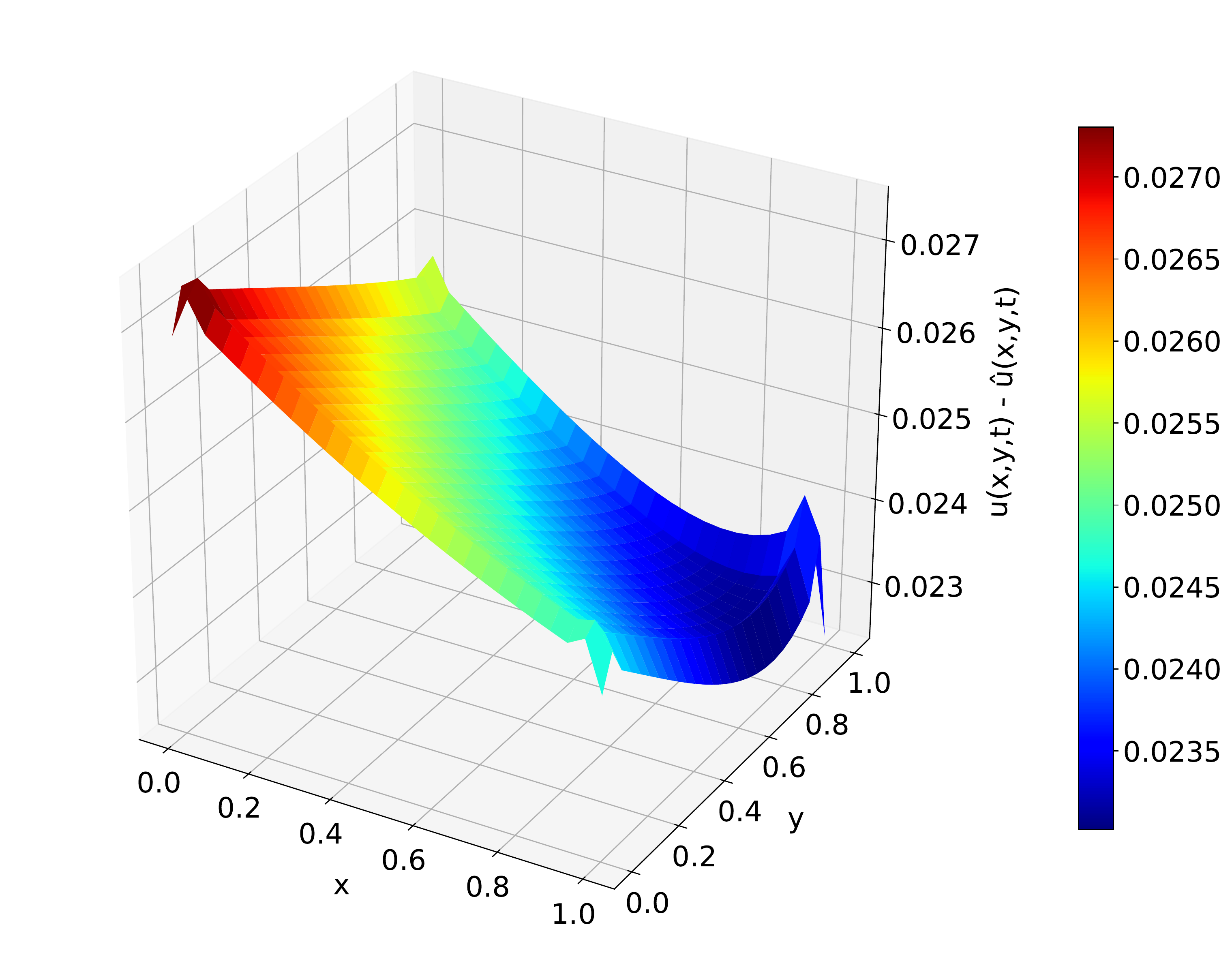}
    \vspace{-.11\textwidth}\caption{}
  \end{subfigure}\hspace{0.02\textwidth}

  \caption{The plots for the 2D-Burgers equation obtained by GCN-FFNN model. (a) and (b): The true solution $u(x,y,t)$. (c) and (d): The obtained residuals $u(x,y,t)-\hat{u}(x,y,t)$. The data at $t=1$ and $t=3$ are from the training set and test set (outside the domain), respectively.}
  
  \label{fig:2d-burgers}
\end{figure*}

Fig. \ref{fig:2d-burgers} (a) and (b) show the true solution at $t=1$ and $t=3$ from training and test set, respectively. The obtained residuals are shown in Fig. \ref{fig:2d-burgers} (c) and (d). 

\subsection{2D-Schrödinger Equation}\label{Burgers Equation}

The 2D-Schrödinger equation with boundary and initial conditions are depicted in Eq. (\ref{eq:2d-schrödinger-1}) \cite{li2021deep}

\begin{equation} \label{eq:2d-schrödinger-1}
\begin{cases}
i\psi_t + \psi_{xx}+\psi_{yy}+w(x,y)\psi =0,\; (x,y) \in [-5,5],\; t\in [0,1], \\
\psi(x,y,t)=ie^{it}/(cosh(x)+cosh(y)) \\

\end{cases}
\end{equation}

\noindent
with
\begin{equation} \label{eq:2d-schrödinger-2}
w(x,y) = 3-2tanh^2(x)-2tanh^2(y), \\
\end{equation}

\noindent
Here, the differential operator is defined as $f[u]:=f=i\psi_t + \psi_{xx}+\psi_{yy}+w(x,y)\psi.$ Let $u$ and $v$ denote the real and imaginary components of $\psi$, then the 2D-Schrödinger equation can be rewritten as follows \cite{li2021deep}:

\begin{equation} \label{eq:2d-schrödinger-4}
\begin{cases}
u_t = -v_{xx}-v_{yy}-wv, \\
v_t = u_{xx}+u_{yy}-wu. \\

\end{cases}
\end{equation}

\begin{figure*}[hbt!]
\center{}
%
  \begin{subfigure}{.33\textwidth}
    \figuretitle{The plots of $u(x,y,t)$ at $t=1$}
    \includegraphics[width=\textwidth]{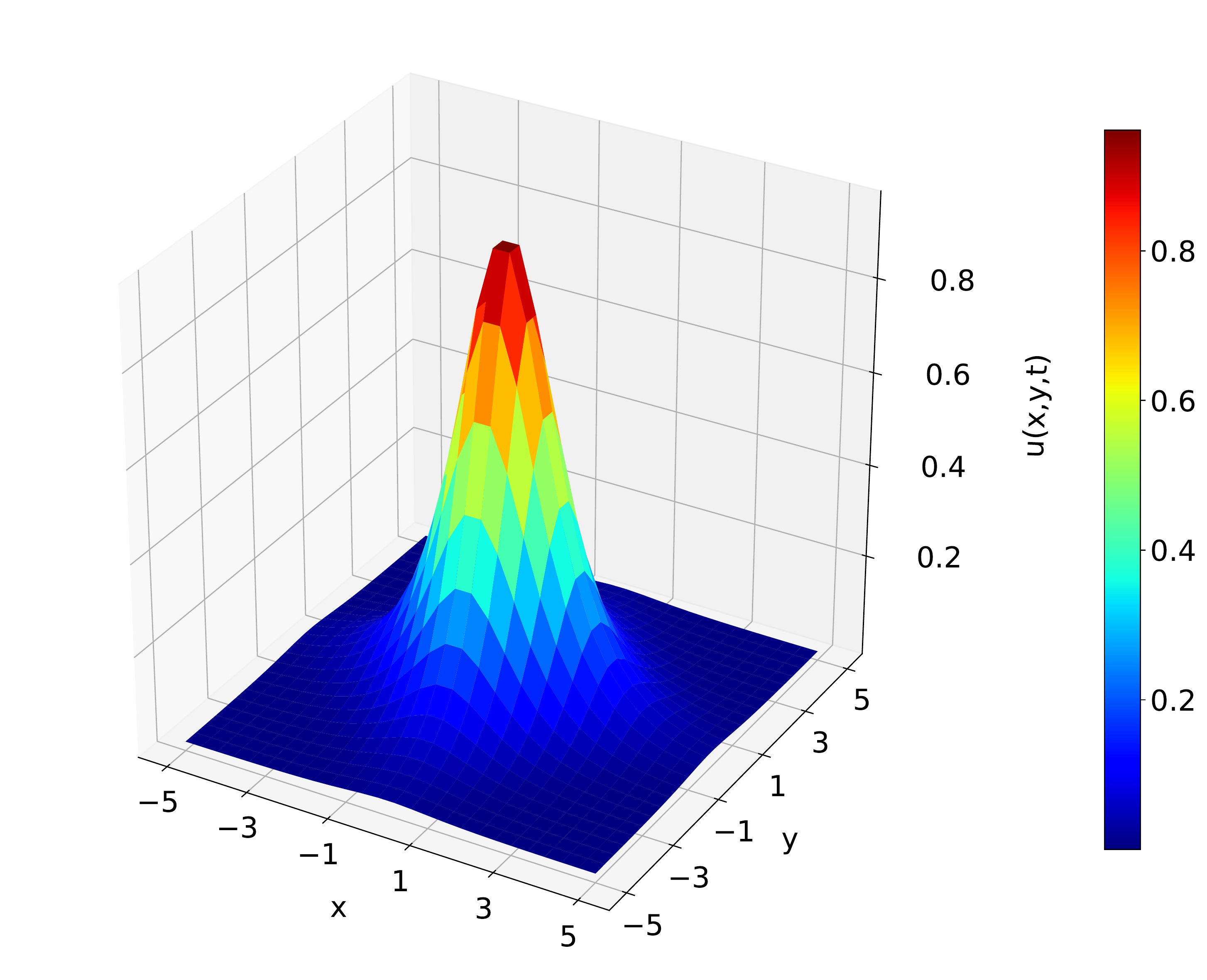}
    \vspace{-.11\textwidth}\caption{}
  \end{subfigure}\hspace{0.0\textwidth}
%
%
%
  \begin{subfigure}{.33\textwidth}
    \figuretitle{The plot of $u(x,y,t)-\hat{u}(x,y,t)$ at $t=1$}
    \includegraphics[width=\textwidth]{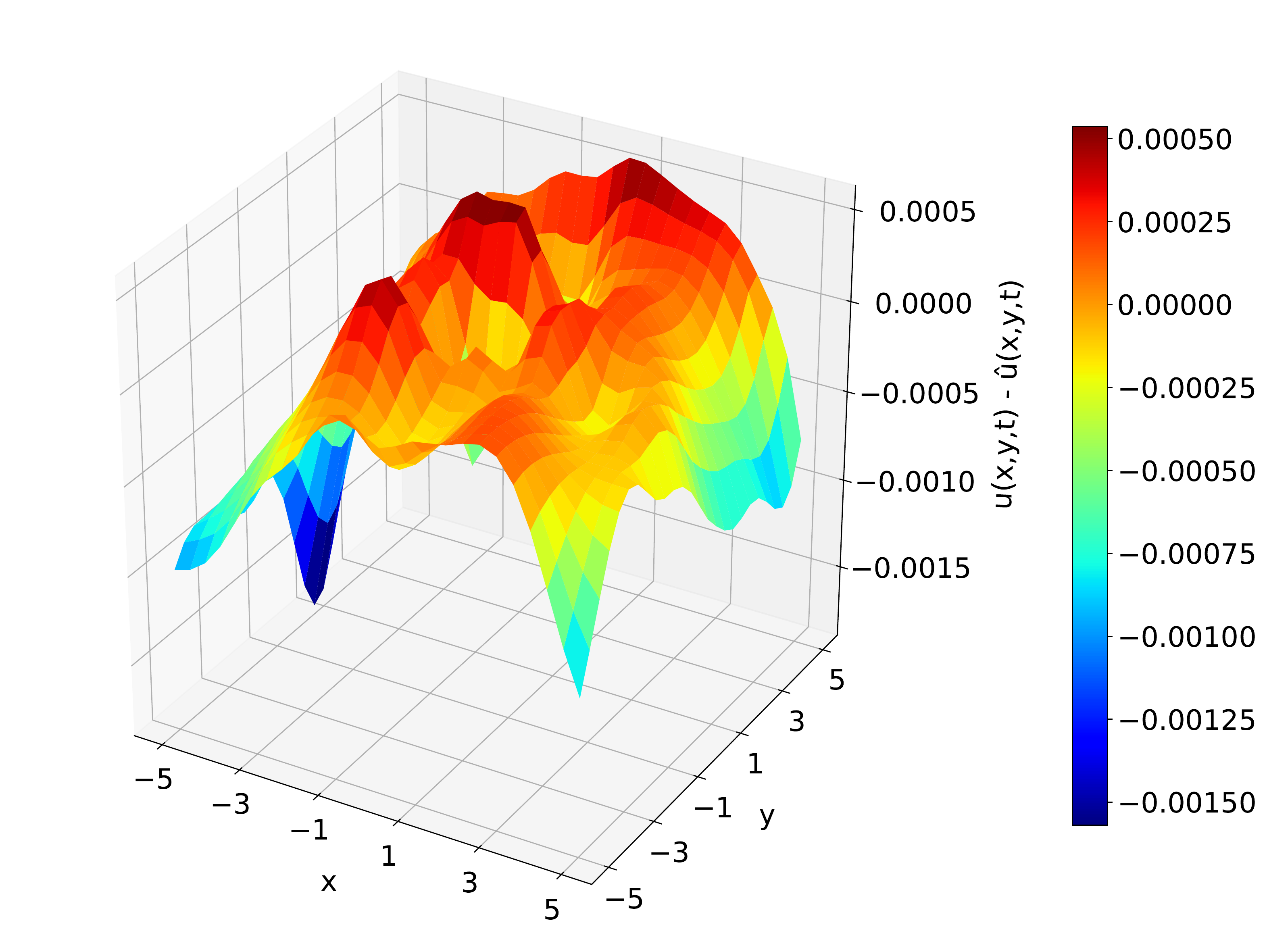}
    \vspace{-.11\textwidth}\caption{}
  \end{subfigure}\hspace{0.0\textwidth}

  \caption{The plots for the 2D-Schrödinger equation obtained by GCN-FFNN model for the outside domain test at $t=1$. (a): The true solution $u(x,y,t)$. (b): The obtained residuals $u(x,y,t)-\hat{u}(x,y,t)$.}
  \label{fig:2d-schrödinger}
\end{figure*}

The domain of the 2D-Schrödinger equation, i.e. $(x,y,t) \in [-5,5]\times[-5,5]\times[0,1]$, is divided into $N=26\times26\times11$ nodes. In the first scenario, $10\%$ of $N$ nodes are randomly selected to form the inside domain test nodes and the remaining nodes are used for training the models. In the second scenario, the nodes in the ranges $(x,y,t) \in [-5,5]\times[-5,5]\times[0,0.9]$ are used for training and the outside domain test nodes are selected from $0.9 < t\leq 1$. As can be seen from Table \ref{tab:loss-inside}, for the 2D-Schrödinger equation, FFNN  and GCN-FFNN models achieved comparable results on inside domain test nodes using both MSE and infinity norm metrics. For outside domain test nodes, the GCN-FFNN model outperforms other models using the MSE metric while it also achieved comparable results to the FFNN model in terms of infinity norm metric. The true solution and the obtained residual at outside domain test time $t=1$ are shown in Fig. \ref{fig:2d-schrödinger} (a) and (b), respectively.


\section{Conclusion}\label{Conclusion}

In this paper, a new two-stream architecture based on graph convolutional network (GCN) and feed-forward neural networks (FFNN) is developed for solving partial differential equations (PDEs). The model learns from both grid and graph input representations obtained by discretizing the domain of the given PDE. The proposed model is examined on four nonlinear PDEs, i.e. 1D-Burgers, 1D-Schrödinger, 2D-Burgers and 2D-Schrödinger equation. The performance of the models are evaluated on test data located inside and outside the domain. Thanks to the incorporation of both types of input representations, most of the time the proposed GCN-FFNN model outperforms the other tested models for the studied PDEs. The implementation of our GCN-FFNN model is available at \footnote{\url{https://github.com/onurbil/pde-gcn}}.

\bibliography{GCN_FFNN}
\end{document}